\documentclass[conference]{IEEEtran}
\usepackage{times}

\usepackage[numbers]{natbib}
\usepackage{multicol}
\usepackage[bookmarks=true]{hyperref}

\usepackage{amsmath,amsfonts}
\usepackage{textcomp}
\usepackage{stfloats}
\usepackage{url}
\usepackage{verbatim}
\usepackage{graphicx}
\usepackage{times}
\usepackage{epsfig}
\usepackage{amssymb}
\usepackage{booktabs}
\usepackage{makecell}
\usepackage{multirow}
\usepackage{tabu}
\usepackage{diagbox}
\usepackage{bbding}
\usepackage{xcolor}
\usepackage{multicol}
\usepackage{color}
\usepackage{bm}
\usepackage{mathrsfs}
\usepackage{pifont}
\usepackage{wasysym}
\usepackage{psfrag}
\usepackage{indentfirst} 
\usepackage{comment}
\usepackage{threeparttable}
\usepackage{colortbl}

\usepackage{algpseudocode}
\usepackage{algorithm}

\IEEEoverridecommandlockouts

\usepackage{subfigure}

\begin{document}

\title{Leveraging Large Language Model for Heterogeneous Ad Hoc Teamwork Collaboration}

\author{\IEEEauthorblockN{Xinzhu Liu$^{1,\dag}$, Peiyan Li$^{1,\dag}$, Wenju Yang$^{2,\dag}$, Di Guo$^{2,*}$, and Huaping Liu$^{1,*}$
\thanks{$^{\dag}$Xinzhu Liu, Peiyan Li and Wenju Yang are with equal contributions.}
\thanks{$^{1}$Xinzhu Liu, Peiyan Li and Huaping Liu are with the Department of Computer Science and Technology, BNRist, Tsinghua University, Beijing.}
\thanks{$^{2}$Wenju Yang and Di Guo are with the School of Artificial Intelligence, Beijing University of Posts and Telecommunications, Beijing, China.}
\thanks{$^{*}$Corresponding authors: Di Guo (guodi.gd@gmail.com) and Huaping Liu (hpliu@tsinghua.edu.cn). This work was jointly supported by the National Natural Science Fund for Distinguished Young Scholars under Grant 62025304 and National Natural Science Foundation Project under Grant 62273054.}
}
}

\maketitle

\begin{abstract}
Compared with the widely investigated homogeneous multi-robot collaboration, heterogeneous robots with different capabilities can provide a more efficient and flexible collaboration for more complex tasks. In this paper, we consider a more challenging heterogeneous ad hoc teamwork collaboration problem where an ad hoc robot joins an existing heterogeneous team for a shared goal. Specifically, the ad hoc robot collaborates with unknown teammates without prior coordination, and it is expected to generate an appropriate cooperation policy to improve the efficiency of the whole team. To solve this challenging problem, we leverage the remarkable potential of the large language model (LLM) to establish a decentralized heterogeneous ad hoc teamwork collaboration framework that focuses on generating reasonable policy for an ad hoc robot to collaborate with original heterogeneous teammates. A training-free hierarchical dynamic planner is developed using the LLM together with the newly proposed Interactive Reflection of Thoughts (IRoT) method for the ad hoc agent to adapt to different teams. We also build a benchmark testing dataset to evaluate the proposed framework in the heterogeneous ad hoc multi-agent tidying-up task. Extensive comparison and ablation experiments are conducted in the benchmark to demonstrate the effectiveness of the proposed framework. We have also employed the proposed framework in physical robots in a real-world scenario. The experimental videos can be found at {\url{https://youtu.be/wHYP5T2WIp0}}.

\end{abstract}

\IEEEpeerreviewmaketitle

\section{Introduction}

\begin{figure}
	\centering
    \setlength{\abovecaptionskip}{-0.5cm}
	\includegraphics[width=3.45in]{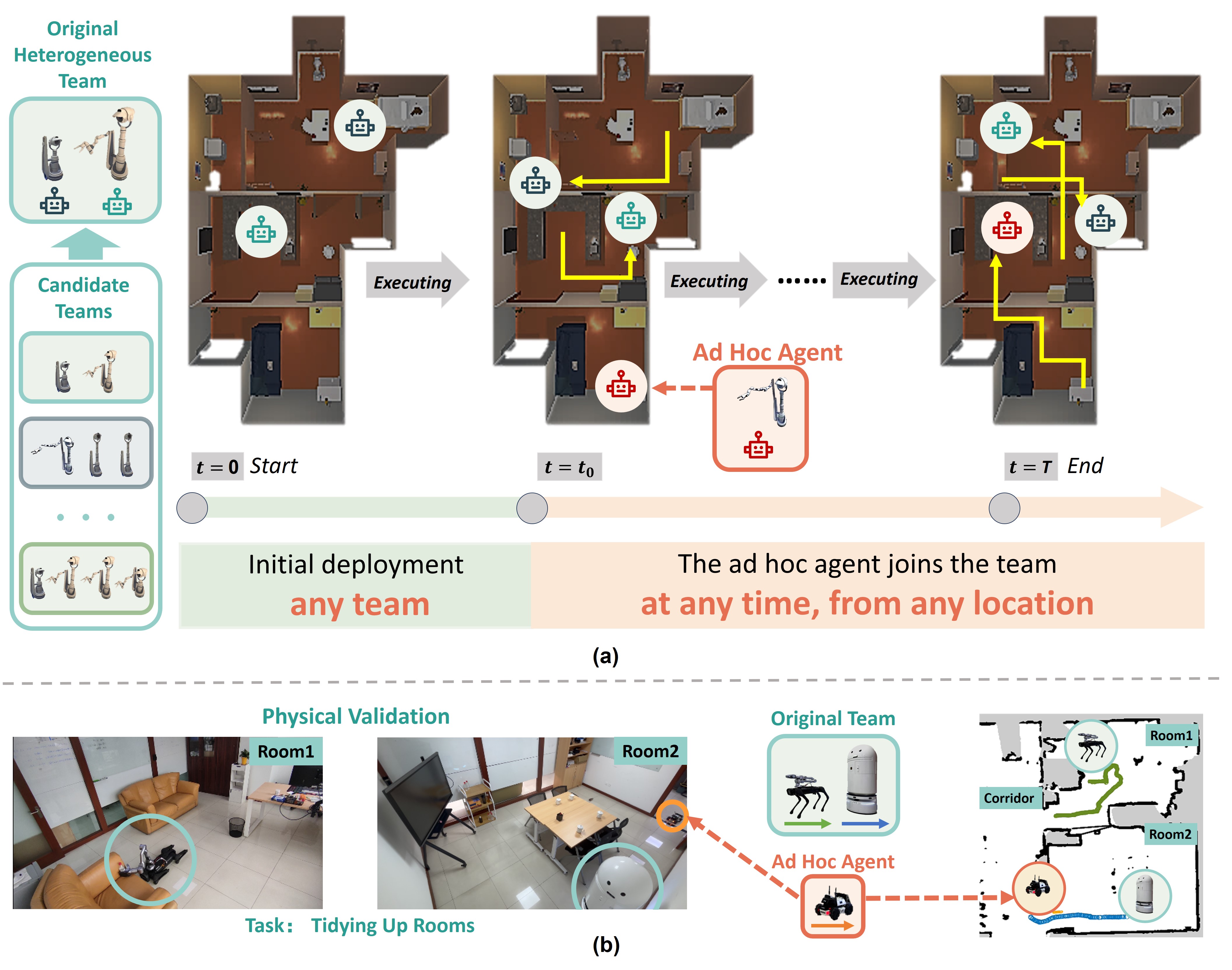}
	\caption{The illustration of heterogeneous ad hoc teamwork collaboration. The heterogeneous ad hoc agent joins any original team at any time from any location, and then the new team collaborates to finish the task. {As illustrated in (a), at the start, an original team is randomly selected from the group of candidate teams to execute a task. When $t = t_0$, an ad hoc agent joins the team from any given location. The ad hoc agent seamlessly integrates into the team without prior coordination. Then, the new team collaborates and finally finishes the task. As illustrated in (b), the proposed framework can be applied to the physical validation experiments with real robots.}}
    \vspace{-1.6em}
	\label{fig:illustration}
\end{figure}

Imagine after a natural disaster such as an earthquake or hurricane, a team of robots is dispatched for the rescue task. Since the situation of a disaster site is complex, robots of different capabilities may be required for the rescue. These robots are likely to be brought from different places and thus arrive at the site at different times. The coming robot doesn't have any prior information on existing teammates, and it is expected to collaborate efficiently and robustly with previously unknown teammates for the same goal. This scenario describes a typical heterogeneous ad hoc teamwork, and the new coming robot is called an ad hoc robot. Specifically, a heterogeneous ad hoc teamwork collaboration is demonstrated in Fig. \ref{fig:illustration}, where heterogeneous robots of different capabilities can compose any team, and the original heterogeneous team collaborates to execute a task. An ad hoc robot could join this team at any time from any location, and then a heterogeneous ad hoc team could collaborate to achieve the goal. 

During the past years, the multi-robot collaboration task has been widely investigated, and a bunch of multi-agent embodied tasks are proposed where multiple agents learn proper strategies to collaborate efficiently \cite{jain2019two,jain2020cordial,liu2022multi,liu2022embodied,tan2020multi,wang2021collaborative,wang2023multi,yu2022learning} and solve complex embodied tasks \cite{miao2023long,saricciccek2023novel}. All these works only consider homogeneous agents with the same capabilities. However, in real-world applications, the robots may be faced with more complicated situations such as seismic sensing and planetary exploration. It is necessary to leverage robots with different capabilities to accomplish the task better \cite{howard2006experiments,ju2019modeling,schuster2020arches,sharma2022ch,sudarshan2017heterogeneous}. Meanwhile, the ad hoc teamwork collaboration is an important problem in the heterogeneous multi-robot collaboration, which has been rarely addressed. In the heterogeneous ad hoc team collaboration task, an ad hoc robot has to collaborate with a group of unknown teammates without any prior coordination. An appropriate cooperation policy should be generated for the ad hoc robot to improve the efficiency of the whole team and avoid interfering with the effective collaboration among original teammates, which makes it an extremely challenging task. 

The formal definition of ad hoc teamwork is proposed in \cite{stone2010ad}, and has attracted a surge of studies in the past years\cite{albrecht2016belief,albrecht2015hba, barrett2017making,chen2020aateam,dodampegamacoordination, dodampegama2023toward,zintgraf2021deep}. However, most of these works discuss ad hoc teamwork collaboration as a theoretical problem under relatively ideal conditions in some simple 2D environments. Ref. \cite{wang2021collaborative} considers the ad hoc setting in the 3D embodied environment. Still, it only discusses the situation that the number of agents in the training process differs from that in the testing process without thinking of an ad hoc agent adapting to changing teams. Furthermore, prior studies mostly focus on the situation in which the ad hoc agent joins the team at the beginning of the task, but in practical scenes, the ad hoc agent is likely to join the team at any time. Therefore, existing ad hoc teamwork studies are difficult to apply directly to practical robotic applications. In this paper, we focus on a more general ad hoc teamwork setting in which the ad hoc agent can join any team, at any time, and from any location.

On the other hand, the Large Language Model (LLM) has recently shown promising performance in natural language understanding, reasoning, task planning, as well as decision-making in robotics \cite{huang2022inner, sarch2023open}. And there emerge a lot of works trying to leverage the LLM to perform high-level planning for robots \cite{chen2023robogpt,huang2022language,huang2022inner,liu2023reflect,prasad2023adapt,qiao2023march,yao2022react}. To generate the reasonable policy for an ad hoc robot to collaborate with a group of unknown teammates without any prior coordination, we propose to leverage the LLM to perform robust high-level reasoning and planning for the ad hoc agent to effectively collaborate with any team at any time. A novel framework is established for heterogeneous ad hoc teamwork collaboration. The main contributions of this paper are as follows: 
\begin{itemize}

\item \textbf{Framework:} We propose a decentralized heterogeneous ad hoc teamwork collaboration framework, which supports the ad hoc agent to join any previously unknown teammates at any time from any location.

\item \textbf{Method:} A hierarchical dynamic planning model based on the LLM is developed, in which the Interactive Reflection of Thought (IRoT) method is proposed for the LLM to generate the sub-task and a sub-skill planner is developed to predict the next sub-skill for the ad hoc agent to execute. This model can adapt to teammates with different capabilities and policies.

\item \textbf{Validation:}  We build a benchmark testing dataset based on ProcTHOR-10K \cite{procthor} to evaluate the proposed ad hoc teamwork collaboration framework in the heterogeneous multi-agent tidying-up task. Experiments are conducted both in the simulation and physical scenarios, illustrating the effectiveness of the proposed heterogeneous ad hoc collaboration framework.
\end{itemize}

\section{Related Work}

\subsection{Heterogeneous Multi-Agent Tasks}
Heterogeneous multi-agent collaboration tasks leverage the diverse capabilities and complementary attributes of the robots to enhance the scalability and robustness of the multi-agent system. There are already some heterogeneous robot combinations proposed such as aerial-ground collaboration \cite{lissandrini2020decentralized, yue2021tightly}, main-picket collaboration \cite{haldane2014detection}, and humanoid-quadruped collaboration \cite{wallace2020multimodal}. However, challenges in coordination, planning, and execution are also presented. Some methodologies that focus on fixed robot combinations have been proposed. Zhao et al. \cite{zhao2022stackelberg} generate coordination strategies for robots with asymmetric information and influence. Decentralized cooperative transportation by aerial and ground robots equipped with manipulators is discussed in \cite{lissandrini2020decentralized}. Haldane et al. \cite{haldane2014detection} undertake joint locomotion and perception tasks by legged robots of varying sizes and capabilities. There is also work dealing with multi-task collaboration scenarios involving a variety of heterogeneous robots and humans \cite{logothetis2021efficient}. Our research focuses on improving collaboration within any team, irrespective of the diversity in capabilities, policies, and team size.

\subsection{Ad Hoc Teamwork}

Ad hoc teamwork involves designing a collaboration policy for agents to collaborate with unknown teammates without prior coordination, which was initially proposed by Stone et al. \cite{stone2010ad} and further detailed by Mirsky et al. \cite{mirsky2022survey}. Early research mostly focuses on matrix games, which assume knowledge of partner actions \cite{albrecht2012comparative,stone2009leading}. Two common methods for ad hoc teamwork include type inference, which models teammates as defined types and uses past interactions to estimate type probabilities for learning algorithms \cite{albrecht2016belief,albrecht2015hba}, and experience recognition, which compares current and past observations to identify the best actions, with PLASTIC-Policy as an influential example \cite{barrett2017making}. Deep learning has been applied to both methods \cite{chen2020aateam,papoudakis2019dealing,zintgraf2021deep}, and some variations of ad hoc teamwork have been explored, including partial observability settings \cite{gu2021online,papoudakis2021agent} and open environment settings with fluctuating team sizes \cite{eck2020scalable,rahman2021towards}. However, most ad hoc teamwork research uses data-driven learning methods, which are computationally expensive and lack transparency. To address this, an architecture is introduced that uses non-monotonic logical reasoning with prior commonsense domain knowledge and models learned from limited examples to predict other agents’ behavior \cite{dodampegamacoordination}. This approach is further developed into a framework that supports non-monotonic logical reasoning with prior commonsense domain knowledge \cite{dodampegama2023toward}. Our work expands existing ad hoc teamwork research by considering settings where agents have different arrival times.

\subsection{LLMs for Embodied Multi-Robot Tasks}
Large Language Models (LLMs) have been increasingly utilized to solve embodied collaboration tasks among multiple robots. In \cite{chen2023scalable}, four distinct communication mechanisms based on LLMs have been proposed, and their efficiency and scalability have been studied in 2D multi-agent tasks. In the Overcooked-AI game, the \textit{ProAgent} leverages LLMs for cooperative reasoning to enhance collaboration \cite{zhang2023proagent}. For 3D embodied visual tasks, LLMs empower embodied agents to communicate, plan, and collaborate with other agents or humans \cite{zhang2023building}. The multi-robot manipulator benchmark, \textit{RoCoBench}, introduces LLMs to facilitate communication, generate subtask plans, and perform multi-arm motion planning \cite{mandi2023roco}. In the context of homogeneous multi-agent visual semantic navigation, \textit{Co-NavGPT} employs LLMs as a central planner to predict the next navigation frontier for each agent \cite{yu2023co}. Also, \textit{SMART-LLM} uses LLMs for central task decomposition and task allocation in fully observable heterogeneous multi-agent tasks, considering the different abilities and states of agents \cite{kannan2023smart}. These works highlight the versatility and potential of LLMs in facilitating effective multi-agent collaboration.

\section{Problem Formulation}
\label{3}
In this work, we aim to address the heterogeneous ad hoc teamwork collaboration problem. We examine a scenario involving an ad hoc agent collaborating with unfamiliar teammates at any time, from any location. In the absence of prior coordination, the agent must adapt to the original team by communicating basic information with teammates and observing teammates' behaviors. 

Given a goal $\mathcal{G}$ and an existing team $\mathcal{T} = \{T_1,\cdots,T_N\}$, where $T_n$ is the $n$-th embodied agent and $N$ is the number of teammates \footnote{{In the task setting of this paper, the teammates are all agents or robots and do not contain humans.}}. We denote the policy set as $\Pi =\{\pi_1, \pi_2, \cdots, \pi_N\}$, where $\pi_n$ is the policy of the $n$-th embodied agent $T_n$. 

Though the multi-agent collaboration is still not fully addressed, there exists a lot of work that provides many feasible solutions. Therefore, in this work we assume we have already obtained a feasible policy set $\bar{\Pi}$ for the existing team, and the objective is to design the policy for an ad hoc agent $T_0$, which may enter the environment at any time-step (denoted as $t_0\geq 0$), to interact with existing agents in $\mathcal{T}$ to complete the goal $\mathcal{G}$. The policy of the ad hoc agent $T_0$ is denoted as $\pi_0$, of which optimal solution can be obtained as  
\begin{equation}\label{op_2}
\pi_0^* = \mathop{\arg\max}\limits_{\pi_0}~~ U_{t\geq t_0}(\mathcal{G}, \mathcal{T}\cup T_0, \bar{\Pi}\cup \pi_0)
\end{equation}
where $U_{t\geq t_0}()$ is the utility function which is dependent on the goal, agents, and policies. The subscript $t\geq t_0$ means it is accumulated from the time instant $t_0$. Note that in the above equation, $\bar{\Pi}$ is the fixed policy for existing teammates, and the utility function could be different from the one that is used to optimize $\bar{\Pi}$. This guarantees the existing team $\mathcal{T}$ is indeed a black box to the ad hoc agent $T_0$.

Note that even though we set $t_0 = 0$, the problem in Eq. \ref{op_2} is totally different from the conventional multi-agent collaboration. Because in our ad hoc setting (Eq. \ref{op_2}), we can only optimize the policy of the ad hoc agent, while other agents keep their policies unchanged. In addition, the ad hoc agent has no prior information about the existing teammates $\mathcal{T}$. During the collaboration procedure, the ad hoc agent should observe the situation, communicate with its teammates, and try to adapt to their behaviors, which is a challenging zero-shot collaboration problem.

Further, we give more detailed descriptions of the embodied agent, which clarify the boundary of the problem in this work. In the following, we do not differentiate the ad hoc agent ($n=0$) and the existing teammates($n=1,2,\cdots, N$).

For the $n$-th embodied agent (for $n=0,1,2,\cdots, N$), we define a 5-tuple $\mathcal{W}_n = \{\mathcal{S}_n, \mathcal{M}_n, \mathcal{C}_n, \mathcal{A}_n, \mathcal{H}_n\}$, which are illustrated as
\begin{enumerate}
    \item $\mathcal{S}_n$: State space, which includes the observations, the status(location, orientation, power assumption, ...), etc
    \item $\mathcal{M}_n$: Messages space, which includes the communication messages with other available agents  
    \item $\mathcal{C}_n$: Capability space, which includes the agent capability of action, payload, and resources  
    \item $\mathcal{A}_n$: Action space, which includes the practically execution actions
    \item $\mathcal{H}_n$: Memory space, which includes historical data containing states, actions, messages, etc
\end{enumerate}

Note that the above five sets are mutually coupled. For example, the action $\mathcal{A}_n$ is obviously influenced by the capability space $\mathcal{C}_n$. In addition, $\mathcal{W}_{n_1}$ and $\mathcal{W}_{n_2}$ can be significantly different for $n_1 \neq n_2$, which is a prerequisite for heterogeneous ad hoc collaboration. Based on the above setting, we formalize the corresponding policy as the mapping 
\begin{equation}\label{op_3}
\pi_n(\cdot): \{\mathcal{S}_n, \mathcal{M}_n, \mathcal{C}_n, \mathcal{H}_n \} \rightarrow \Delta(\mathcal{A}_n)
\end{equation}
where $\Delta$ represents probability distribution over the corresponding set.

Before closing this section, we emphasize that the above problem setting strictly complies with the assumptions summarized in \cite{mirsky2022survey}: (1)No prior coordination. The ad hoc agent is expected to cooperate with its teammates when the task begins without any prior opportunities to establish or specify mechanisms for coordination. (2) No control over teammates: The ad hoc agent cannot change the properties of the environment, and the teammates’ policies and communication protocols, it has to reason and act under the given conditions. {The properties of the environment include the observability level, the modality of the observation information the agent obtains, the categories of objects that may exist in the environment, and so on, which is different from the state of the environment.}

\section{Methodology of ad hoc Teamwork}
\begin{figure*}
	\centering
	\includegraphics[width=7in]{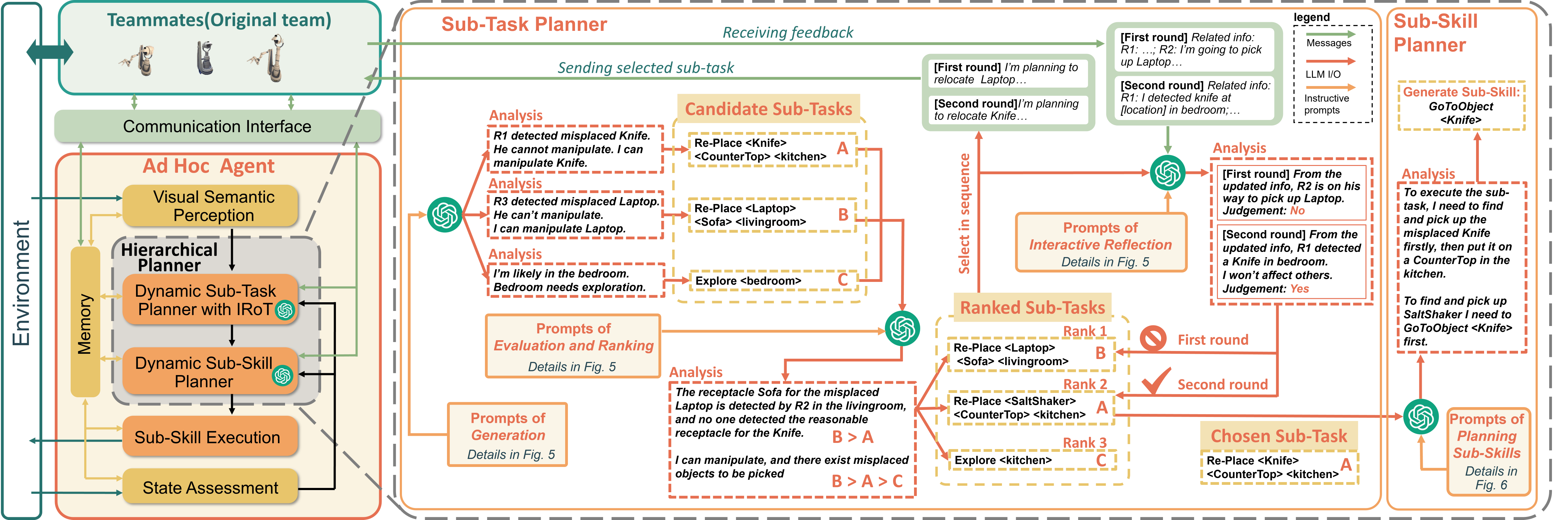}
	\caption{The overview of the proposed ad hoc teamwork framework. The ad hoc agent exchanges effective information with the original team through the communication interface. The visual semantic perception takes the RGB-D and poses as input to build the top-down semantic map and the hierarchical scene graph. The dynamic sub-task planner based on the LLM utilizes the newly proposed IRoT method to generate the next sub-task intention for the ad hoc agent. The dynamic sub-skill planner predicts the next sub-skill for the ad hoc agent. Then, low-level actions are generated to execute the planned sub-skill. The state assessment module evaluates the new state after the execution of low-level actions in the corresponding sub-skill and generates effective information to help re-plan the sub-task or sub-skill. {The left part shows the overall structure. The right part takes a closer look into the hierarchical planner. The entire planning process of the sub-task planner is illustrated. The sub-task planner with the LLM generates three candidate sub-tasks (A, B, and C) in the generation stage and ranks them as B, A, and C in the evaluation stage. Two rounds judgement in the interactive reflection stage are shown in the right part. The selected sub-task B is judged to be infeasible in the first round. Then the sub-task A is judged to be reasonable in the second round and chosen as the current planned sub-task. The sub-skill planner takes the chosen sub-task as input and generate the analysis and planned sub-skill with the LLM.}
 }
	\label{fig:model}
	\vspace{-1.0em}
\end{figure*}

An overview of the proposed decentralized heterogeneous ad hoc teamwork collaboration framework is demonstrated in Fig. \ref{fig:model}, which plans and generates the action decision only with the state information of the ad hoc agent and the communication messages from teammates. The decentralized framework means that there does not exist a central node to generate the action decision for all teammates and the ad hoc agent by collecting all their state information. The visual semantic perception module is responsible for generating the top-down semantic map and the hierarchical scene graph of the environment. An LLM-based hierarchical planner, which consists of a dynamic sub-task planner and dynamic sub-skill planner, is used for generating a coordination policy for the ad hoc agent. Specifically, the dynamic sub-task planner generates the next sub-task for the ad hoc agent to perform with a newly proposed Interactive Reflection of Thought (IRoT) mechanism. Then, the dynamic sub-skill planner generates the next sub-skill to execute based on the generated sub-task, and the sub-skill execution module generates the corresponding low-level actions for the ad hoc agent to execute the sub-skill. Finally, the state assessment module evaluates the action state after the execution and generates effective information to help re-plan the sub-task and sub-skill. The ad hoc agent transmits and receives the information with the original team through the communication module.

We employ the proposed ad hoc framework in the tidying-up task setting, in which multiple agents need to find all misplaced objects and put them in proper receptacles. Specifically, receptacles refer to objects that other objects can be placed on their surface or inside. The misplaced object refers to the object that is put in the unreasonable receptacle according to common sense. In the tidying-up task, agents need to infer whether the detected object is misplaced and infer the reasonable receptacle to place it. Eventually, agents pick up all misplaced objects and put them in reasonable locations.

\subsection{Preliminary Information}
For convenience, we describe the needed information in the proposed framework under the tidying-up setting in advance as follows

\begin{itemize}

    \item State information ($\mathcal{S}_0$): the location of the agent, the object-grasping state, the observation, and key detection information, as well as the task-relevant state in the current time step.
    \item Message information ($\mathcal{M}_0$): all messages the agent has sent and received in the current time step.
    \item Capability information ($\mathcal{C}_0$): agent height, manipulation ability, payload (the maximum mass of objects that can be picked up), and total number of steps the battery power can initially support.
    \item Action information ($\mathcal{A}_0$): the action the agent takes in the current time step.
    \item Memory information ($\mathcal{H}_0$): all state, message, and action information in all previous time steps. 
\end{itemize}

\begin{figure}
    \centering
    \setlength{\abovecaptionskip}{-0.5cm}
	\includegraphics[width=3.45in]{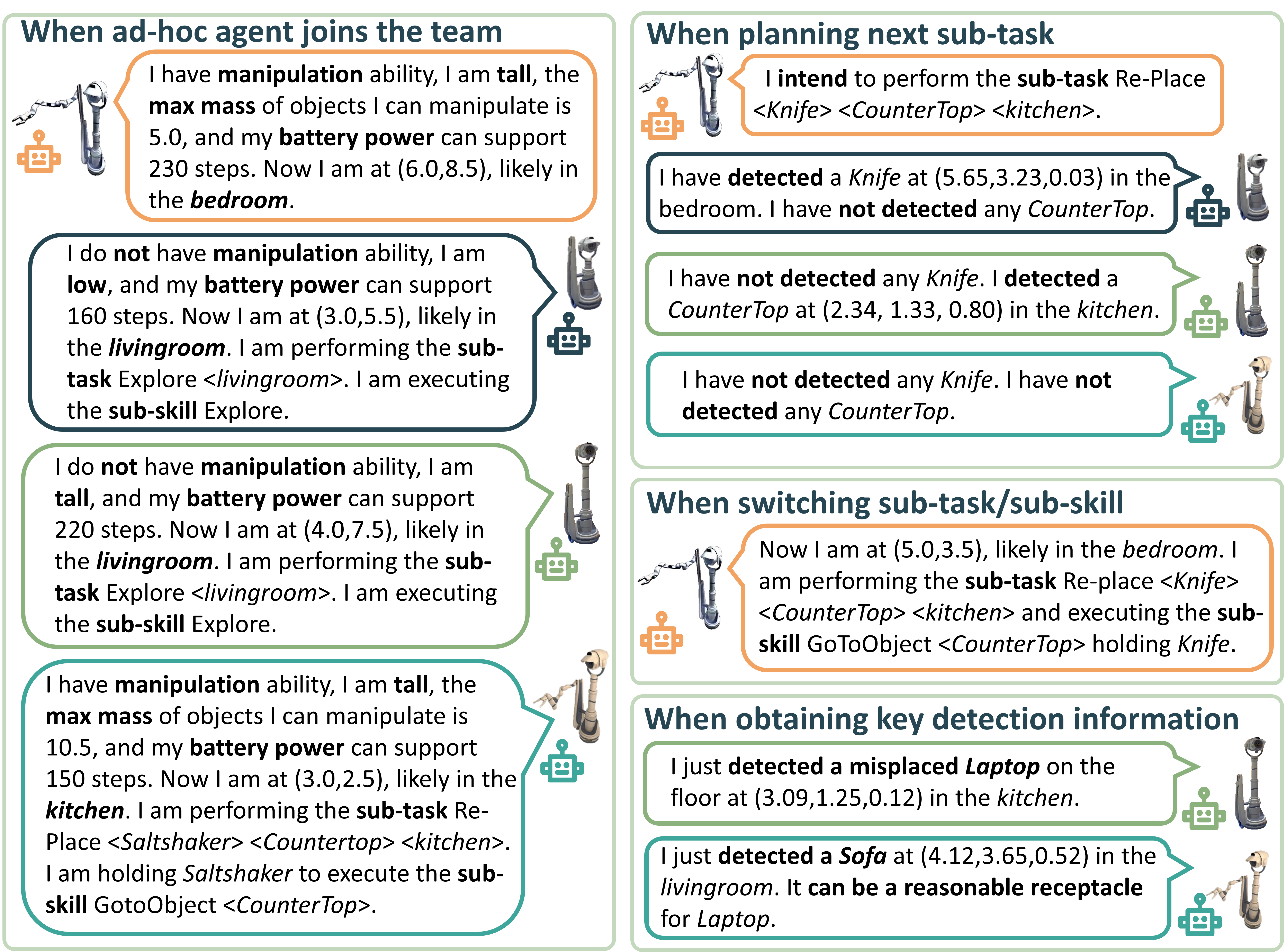}
	\caption{The examples of communication messages.}
	\label{fig:com}
    \vspace{-1.8em}
\end{figure}

\subsection{Communication}

The ad hoc agent utilizes the communication module to exchange effective information with the original team. When the ad hoc agent joins the original team, it will broadcast its state and capability information to teammates and receive reflected messages. {The reflected messages refer to the state and capability information of teammates that are transmitted from each teammate to the ad hoc agent.} Team members are expected to transmit the corresponding messages back. When planning the next sub-task, the ad hoc will broadcast its intention and utilize the received messages from teammates to complete the planning. The task-relevant state contains the current sub-task and sub-skill planning results. When the ad hoc agent has a new task-relevant state, specifically when it finishes or switches to new sub-tasks or sub-skills, it will transmit the information to teammates. The ad hoc agent can also receive the corresponding information from other teammates to do subsequent planning. After execution of each step, the ad hoc agent can reason and judge whether the currently detected object is misplaced and if it can be the receptacle for other misplaced objects. The key detection information refers to these reasoning results. When the ad hoc agent obtains the new key detection information, it will broadcast this information to other teammates. The examples of the above communication are shown in Fig. \ref{fig:com}. We have created an interface for different policies of teams to adapt to the ad hoc agent. This interface extracts capability and state information from the message format of teammates' models, accommodating their varying action policies.

\begin{figure}
    \centering
    \setlength{\abovecaptionskip}{-0.50cm}
	\includegraphics[width=3.45in]{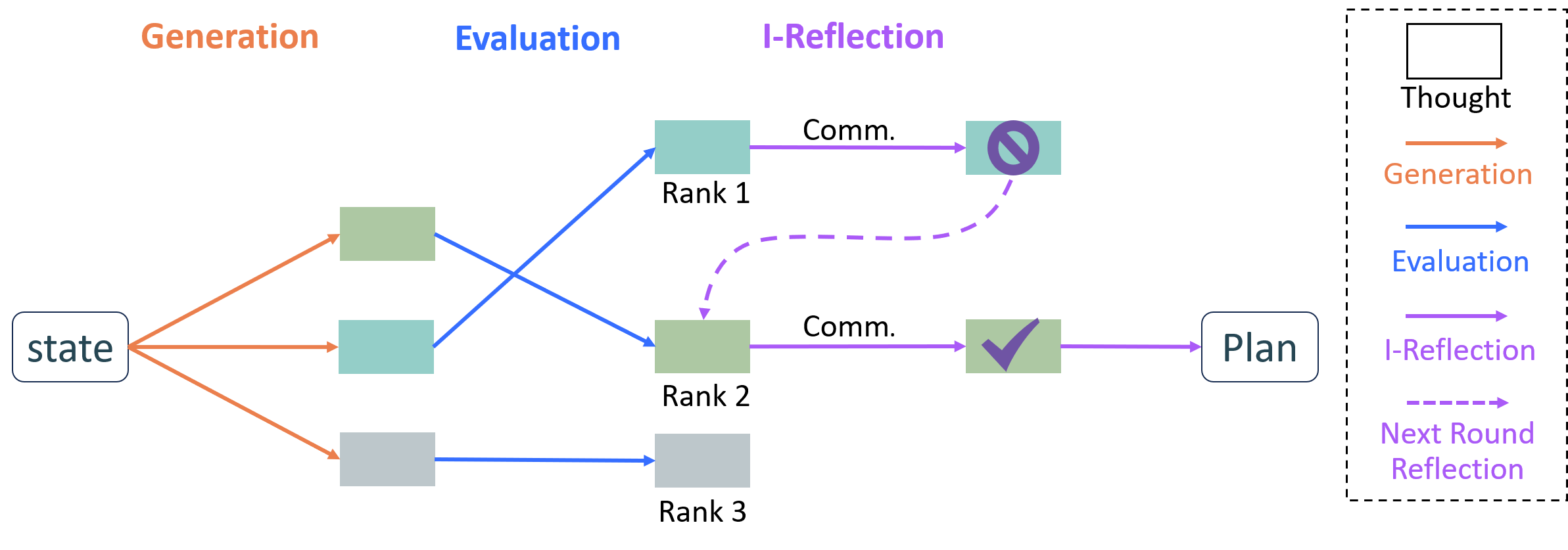}
    \caption{{The structure and the planning process of IRoT.} Each rectangle box represents a \textit{thought}, which is an intermediate sub-task plan toward problem-solving. {Different colors of rectangle boxes signify distinct generated sub-tasks. Lines in different colors represent three stages in the IRoT model. Dotted lines indicate that the chosen sub-task is evaluated to be infeasible and the next chosen sub-task is evaluated with the communication messages. From left to right, $n_{irot}$ sub-task plans are initially generated in the first stage \textit{Generation of Candidate Sub-tasks}. Then these sub-tasks are evaluated and ranked in the second stage \textit{Evaluation \& Ranking}. Finally, IRoT re-judges whether each ranked sub-task is feasible in sequence based on the communication information with teammates in the third stage \textit{Interactive Reflection}. As shown in this figure, the first selected sub-task (\textit{Rank 1}) is evaluated infeasible and the second selected one (\textit{Rank 2}) is evaluated reasonable to execute. IRoT finally selects the second sub-task to execute.}}
	\label{fig:irot}
    \vspace{-0.9em}
\end{figure}

\subsection{Visual Semantic Perception}
The visual semantic perception module detects the objects in the current view of the ad hoc agent, generates the room-object aware semantic map, and obtains the spatial relationships among all detected objects. At each step, the ad hoc agent generates the semantic point clouds based on the obtained RGB-D image, instance segmentation result, and the current pose. The type of room where the agent is located is reasoned according to the categories of currently detected objects. The semantic point clouds are voxelized and flattened to generate a top-down object-room-aware semantic map, which is utilized for exploration and navigation. The hierarchical spatial scene graph is constructed based on the coordinates of each detected object and receptacle as well as the predicted room location, which is utilized to plan the subsequent sub-task and sub-skill. The ad hoc agent updates the semantic map and scene graph with new detection information at each step.

\subsection{Dynamic Sub-Task Planner with IRoT}
\label{subTasKPlanner}

\begin{figure}
    \centering
    \setlength{\abovecaptionskip}{-0.5cm}
    \includegraphics[width=3.45in]{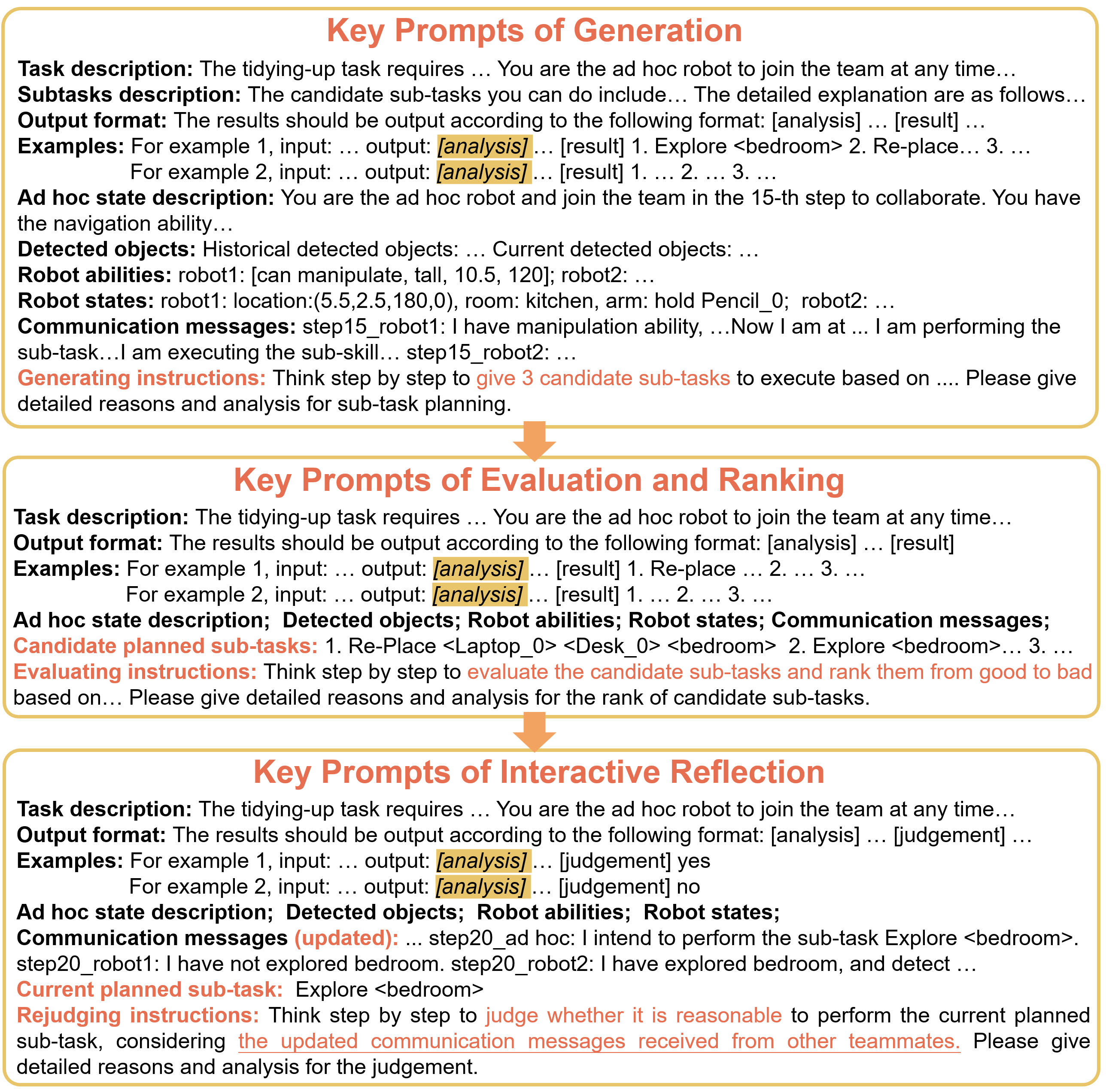}
	\caption{Key prompts of the dynamic sub-task planner with IRoT. {The key prompts of three stages in IRoT are demonstrated respectively, providing additional context to complement Fig. \ref{fig:model}. }}
	\label{fig:Subtask}
    \vspace{-1.5em}
\end{figure}

The LLM is employed as the dynamic sub-task planner to predict the subsequent sub-task for the ad hoc agent. This sub-task is structured to encapsulate the agent's overall intention within a specific time frame. The LLM is utilized for planning due to its ability to analyze and rationalize the sub-tasks that need to be executed based on teammates' capabilities and state information. It also offers flexibility in collaborating with teammates possessing varied skill sets. When joining and collaborating with the original team, the ad hoc agent is likely to have multiple options of sub-tasks to execute when planning, and performing different sub-tasks may have different effects on the original team. The ad hoc agent needs to find the most suitable sub-task to make the collaboration more efficient. Due to the limited information the ad hoc agent obtains when joining the team, the communication messages from teammates are beneficial for the ad hoc agent to make decisions. To solve this complex situation in the interactive environment, we propose the new method \textit{Interactive Reflection of Thoughts (IRoT)} on the basis of the LLM, which introduces the communication reflection mechanism to evaluate the generated candidate sub-tasks with obtained communication messages and determine the next sub-task for the ad hoc agent.

In the context of the heterogeneous multi-agent tidying-up task, the sub-tasks are categorized into two types:

\begin{itemize}
    \item Explore $\left\langle ID_{room} \right\rangle$: Explore the room $ID_{room}$ to search for misplaced objects and obtain information of the room.
    \item Re-place $\left\langle ID_{object} \right\rangle$ $\left\langle Type_{receptacle} \right\rangle$ $\left\langle Type_{room} \right\rangle$: Put the detected misplaced object $ID_{object}$ to the reasonable type of receptacle $Type_{receptacle}$ in the reasonable type of room $Type_{room}$. 
    \item Stop: Stop at the current location.
\end{itemize}

This approach contains three stages, which are demonstrated in Fig. \ref{fig:irot}. The first stage is \textit{Generation of Candidate Sub-Tasks}, in which the LLM generates $n_{irot}$ candidate sub-tasks and the corresponding reasoning analysis with the provided prompts. The second stage is \textit{Evaluation and Ranking}, in which the LLM evaluates the generated sub-tasks and ranks them based on the execution rationality with the candidate sub-tasks and evaluation prompts as input. The third stage is \textit{Interactive Reflection}. In this stage, ranked sub-tasks are sequentially selected and shared with teammates through communication as the ad hoc agent’s intention. The LLM judges the feasibility of the selected sub-task based on the received feedback from teammates. If the selected one is judged to be unreasonable, the next sub-task is selected for another round of this stage. If the selected one is reasonable, the selected sub-task is the planning results of the sub-task planner. The key prompts of the IRoT are shown in Fig. \ref{fig:Subtask}, and the output example is shown in the Appendix.

\subsubsection{\textbf{Generation of Candidate Sub-Tasks}}
We first generate $n_{irot}$ candidate sub-tasks with the Chain-of-Thought (CoT) \cite{wei2022chain} method of the LLM. The state, capability, communication, and memory information are provided to the LLM. It should be noted that all agents obtain partial observations in the environment, which means they can only obtain information about detected objects from their first-person views but cannot get information about all objects in the scene. The action strategies and observation information of other teammates are unknown to the ad hoc agent. The capabilities and executed sub-tasks of each teammate can be extracted from the communication messages.  We give several examples of generation results in the prompt to let the LLM give $n_{irot}$ different candidate sub-tasks while avoiding generating duplicated sub-tasks.

\subsubsection{\textbf{Evaluation and Ranking}}
We utilize the LLM to evaluate and rank the $n_{irot}$ generated candidate sub-tasks with the CoT method. We provide the state, capability, communication, and memory information to the LLM. The LLM takes this information and in-context examples to reason the possible future states after executing different candidate sub-tasks and evaluate the effect on the collaboration with teammates. In this planning process, the LLM considers more details, including the capabilities of the ad hoc agent and teammates, the process of the task execution, and the distance to the detected misplaced objects to rank the cooperation effect of candidate sub-tasks. It outputs the ranked results of $n_{irot}$ candidate sub-tasks according to their rationality in cooperation with teammates.

\subsubsection{\textbf{Interactive Reflection}}
We choose one candidate sub-task from the ranked list of candidate sub-tasks in sequence. The ad hoc agent broadcasts the selected sub-task to other teammates through communication messages. Because the ad hoc agent is likely to join the team after several steps, it does not have the historical state information of teammates. Once the ad hoc agent shares the chosen sub-task, teammates are expected to give responses containing information on previously detected objects or explored rooms. This assists the ad hoc agent in determining the feasibility of the selected sub-task. The LLM uses updated communication and state information to assess the viability of the current sub-task of collaborating with teammates. If the evaluation result of the currently selected sub-task is reasonable, the ad hoc agent will perform this sub-task and make subsequent sub-skill planning based on this sub-task. If the evaluation result is unreasonable, the method will select the next generated sub-task from the ranked list and evaluate it in the same way as above. If the evaluation results of all $n_{irot}$ candidate sub-tasks are not reasonable, the method will re-plan the sub-task from the first step with the newly obtained state information of the task.

The hallucinations of the sub-task planner refer to the generated unreasonable plans of sub-tasks but are deemed syntactically plausible. The proposed IRoT mechanism can detect the hallucinations of the sub-task planner, self-reflects the generated sub-task and modify the generated plans to avoid generating unreasonable sub-tasks with communication information. Only the sub-task that is evaluated to be reasonable in the current task can be executed subsequently.

We have noticed that there is a concurrent work that introduces the cross-model communication to solve the reasoning problem \cite{yin2023exchange}. It mainly pays attention to leveraging different communication structures among multiple LLM models to improve the reasoning result, while our work focuses on solving ad hoc teamwork with the interactive reflection mechanism to collaborate with teammates of different policies, which are not limited to the LLM-based policy.

\subsection{Dynamic Sub-skill Planner}
\label{SubSkillPlanner}
The dynamic sub-skill planner generates the next sub-skill with LLM based on the CoT method for the ad hoc agent. The execution of a sub-task can be decomposed into a list of sub-skills within a certain time sequence. The candidate sub-skills include \textit{GoToObject $\left\langle ID_{object} \right\rangle$}, \textit{GoToPoint $\left\langle \Delta x,\Delta y \right\rangle$}, \textit{GoToRoom $\left\langle ID_{room} \right\rangle$}, \textit{PickupObject $\left\langle ID_{object} \right\rangle$}, \textit{PutObject $\left\langle ID_{object} \right\rangle$ $\left\langle ID_{receptacle} \right\rangle$ $\left\langle ID_{room} \right\rangle$}, \textit{Explore}, \textit{Stop}. The sub-skills are the meta skills that can be executed by the controller. A detailed description of these sub-skills is illustrated in Appendix. We provide the state, capability, communication, and memory information, as well as the in-context examples in the prompts to the LLM. Specifically, the current sub-task and sub-skill plan of each teammate can be obtained from the state and memory information. The key prompts of planning sub-skills are shown in Fig. \ref{fig:Subskill}, and the output example is shown in Appendix.

\begin{figure}
    \setlength{\abovecaptionskip}{-0.5cm}
    \centering
    \includegraphics[width=3.45in]{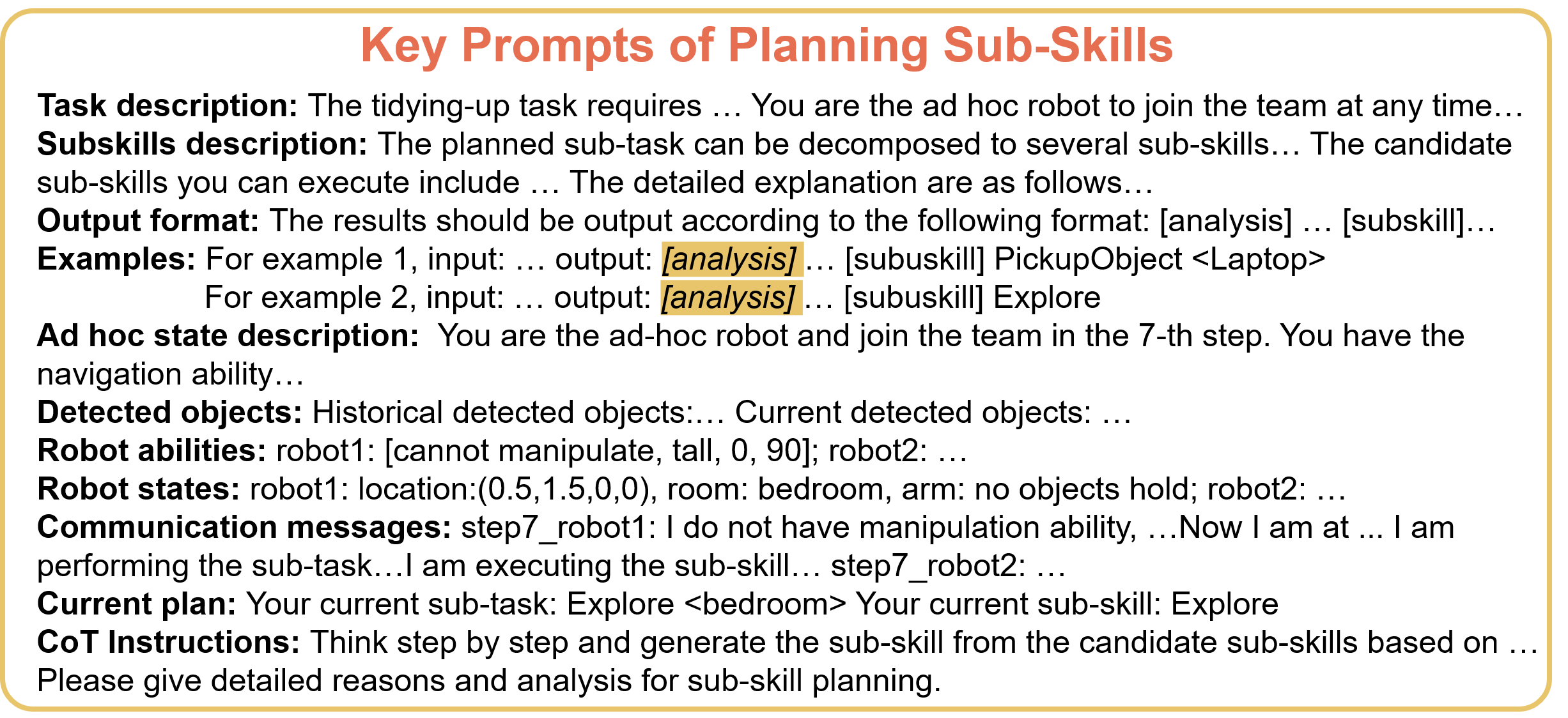}
	\caption{Key prompts of the dynamic sub-skill planner, {providing additional context to complement Fig. \ref{fig:model}.}}
	\label{fig:Subskill}
    \vspace{-1.5em}
\end{figure}

\subsection{Sub-skill Execution}
The sub-skill execution module generates the low-level actions to execute the sub-skill. This module executes the planned sub-skill and updates the ever-changing state information, informing both the sub-task and sub-skill planners of necessary adjustments. We can obtain the execution policies for each sub-skill in advance. The execution policies can be generated through rule-based methods or trained by imitation learning (IL) and reinforcement learning (RL).

\subsection{State Assessment}

{The state assessment module evaluates the success of an action and the state information after executing the corresponding low-level action. The evaluation results would provide more effective information and are likely to lead to the re-planning of sub-tasks or sub-skills. When evaluating the success of the action execution, this module considers whether the ability of the ad hoc agent can match the executed action and whether the action can successfully interact with the environment. For example, the \textit{Pickup} action would fail if the ad hoc agent lacks the manipulation ability, the mass of the object exceeds the maximum mass that the agent can pick, or the agent is far away from the target object when executing the \textit{Pickup} action. If the agent encounters obstacles when performing the navigation action, the action execution would also fail. If the action execution fails, this module would identify the failure reason and provide feedback to the planner and the action execution module to re-generate the reasonable sub-skills and feasible actions. Meanwhile, the agent also judges whether newly detected objects are misplaced or likely to be the target receptacles. If misplaced objects are found, the ad hoc agent would inform its teammates through communication and re-plan the sub-task as well as the sub-skill to execute in the next step. If potential receptacles are identified, the ad hoc agent would share the corresponding information with teammates through communication and re-plan the sub-skill in the next step.}

The hallucinations of the sub-skill planner refer to the generated unreasonable plans of sub-skills but are deemed syntactically plausible. The state assessment module evaluates the action execution state, which can detect the hallucinations of the sub-skill planner and help re-plan the next sub-task as well as the sub-skill.

\subsection{Generalization to Different Tasks}

The proposed framework with the LLM-based hierarchical planner can also be applied to other different tasks. When applying our model to different tasks, the structure of the model remains unchanged. The dynamic sub-task planner with IRoT and the dynamic sub-skill planner are still utilized to generate the next sub-task and the sub-skill to execute. To adapt to different tasks, it is required to design corresponding reasonable candidate sub-tasks and sub-skills for the task, update the description of the new task, modify the descriptions of sub-tasks and sub-skills in the corresponding prompts, and keep the structure as well as other contents in the prompts of the sub-task and sub-skill planner unchanged.

As for the sub-task planner, the whole process of IRoT is kept unchanged and the candidate sub-tasks are designed according to the nature of the corresponding tasks. Then it is necessary to update the description of the new task and modify the description of candidate sub-tasks in the prompts of the sub-task planner as shown in Fig. \ref{fig:Subtask} to the description corresponding to the newly defined sub-tasks. The other contents in the prompts of the sub-task planner can be kept unchanged. The modified model can generate corresponding sub-tasks in new tasks. Meanwhile, as for more complex tasks, if all possible candidate sub-tasks cannot be completely pre-defined, we can remove the sub-task descriptions in the prompts without setting the sub-task types in advance and let the LLM directly generate reasonable sub-tasks according to current information. Then the generated sub-task is input to the sub-skill planner to generate the corresponding sub-skills. The candidate sub-skills are pre-defined and the output of the sub-skill planner must come from the pre-defined types of sub-skills. The structure and other contents in the prompts of the sub-task planner can be kept unchanged.

The sub-skill planner generates the basic skill for the agent to execute according to the current sub-task. In common indoor tasks such as visual navigation \cite{yang2018visual}, visual rearrangement \cite{weihs2021visual}, and ALFRED \cite{shridhar2020alfred}, the manipulation and interaction on objects can be eventually decomposed into the sub-skills of navigation, picking and placing, so the seven candidate sub-skills proposed in our model can cover basic skills in common tasks. When applying our model to different tasks, the task description needs to be updated and other contents in the prompts of the sub-skill planner can remain unchanged. If more complex tasks require other sub-skills to be executed, it is required to design new sub-skills and provide the low-level action strategy under the corresponding sub-skills. Then the sub-skill descriptions in the prompts as shown in Fig. \ref{fig:Subskill} need to be modified to the descriptions of newly defined sub-skills. Other contents in the prompts can be kept unchanged in the sub-skill planner. The sub-skill planner can generate the next sub-skill to execute from the candidate sub-skills.

To apply our model to other tasks, we only need to design the candidate sub-tasks and sub-skills according to the properties of the new task, update the description of the new task, and modify descriptions of the sub-task as well as the sub-skill in the prompts of the sub-task planner and the sub-skill planner.

\begin{figure}
    \centering
    \setlength{\abovecaptionskip}{-0.5cm}
	\includegraphics[width=3.45in]{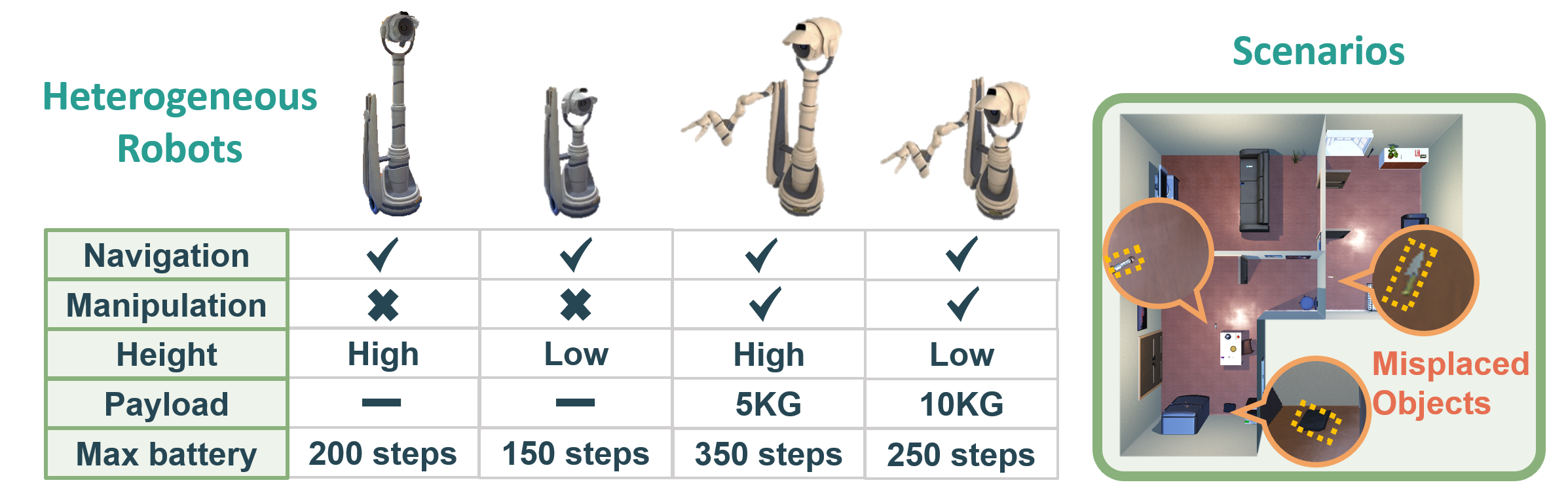}
	\caption{The heterogeneous agents and scene settings in the benchmark dataset.}
	\label{fig:dataset}
    \vspace{-1.8em}
\end{figure}

\section{Experimental evaluation}

\subsection{Dataset}

As far as we know, there isn't any existing benchmark dataset that can evaluate heterogeneous ad hoc teamwork. Considering the tidying-up is an ideal task to evaluate the performance of heterogeneous ad hoc teamwork, we build a benchmark testing dataset for the heterogeneous ad hoc teamwork on the tidying-up task with the ProcTHOR-10K \cite{procthor} dataset.

\textit{Scenario generation}: We randomly select 90 houses from the testing houses in ProcTHOR-10K for evaluation experiments, among which every 10 houses are with 1-8, and 10 rooms respectively (there is no house with 9 rooms in ProcTHOR-10K). We generate 5 tidying-up scenarios for each house, where $k$ pickupable objects are randomly selected and misplaced to unreasonable receptacles or dropped in a random location in the house ($k=1,2,3,4,5$).

\renewcommand{\arraystretch}{0.8}
\begin{table}[!t] 
	\setlength{\abovecaptionskip}{-0.15cm}
	\setlength{\tabcolsep}{0.60mm}
	
	\fontsize{6.8}{9.0pt}\selectfont
	\caption{Quantitative Results in Ad hoc Teamwork with Different Team Policies when \textcolor{red}{$t_0=0$}}
	\label{quanres1}
	\centering
	
	\begin{tabular}{c|c|c|c|c|c|c}
		\Xhline{0.6pt}
		
		\multirow{1}{*}{Team} & \multirow{1}{*}{ad hoc} & \multicolumn{1}{c|}{$\%Suc(\uparrow)$} & \multicolumn{1}{c|}{$\%PS(\uparrow)$}& \multicolumn{1}{c|}{$\#TS(\downarrow)$}  & \multicolumn{1}{c|}{$\#AS(\downarrow)$} & $\%SE(\uparrow)$ \\
		
		
		\Xhline{0.3pt}

        
        \multirow{8}{*}{\makecell{Heuristic \\ based}} & No ad hoc  & 15.3 (-)  & 21.7 (-) & 467.3 (-) & 1863.2 (-) & - \\ 
        \cline{2-7} 
		
		& {Random}  &  15.3 (0) & 21.7 (0)  &  467.3 (0) & 2213 (-18.8) &  0 \\
		& {Heuristic}  & 18.1 (18.3) & 27.4 (26.3) & 435.6 (6.8) & 2049.3 (-10.0) & 1.2  \\
		& {PLASTIC} & 15.6 (2.0) & 22.3 (2.8)  & 456.1 (2.4) & 2165.9 (-16.2) & 0.4  \\
        & {Naive-LLM} & 18.3 (19.6) & 27.9 (28.6) & 436.2 (6.7) & 2060.1 (-10.6) & 1.2  \\
        & {GPL-LLM} & 18.5 (20.9) & 27.9 (28.6) & 433.9 (7.1) & 2046.1 (-9.8) & 1.3 \\
        & {CoT-LLM} & 19.9 (30.0) & 28.5 (31.3) & 421.3 (9.8) & 1987.6 (-6.7) & 2.0  \\
		& {IRoT-LLM} & \textbf{21.0 (37.3)} & \textbf{30.7 (41.5)} & \textbf{410.8 (12.1)} & \textbf{1922.3 (-3.2)} & \textbf{2.6} \\
        \hline\hline

        \multirow{8}{*}{\makecell{Learning \\ based}} & No ad hoc   & 13.5 (-)  & 20.8 (-)  & 473.8 (-) & 1890.2 (-) & - \\ 
        \cline{2-7} 
		
		& {Random} & 13.5 (0) & 20.8 (0)  &  473.8 (0) & 2212.9 (-17.1) & 0 \\
		& {Heuristic}  & 15.6 (15.6) &  25.1 (20.7) &  453.1 (4.4) & 2133.1 (-12.9)& 0.6 \\
		& {PLASTIC} & 13.8 (2.2) & 22.6 (8.7) & 463.8 (2.1) & 2192.1 (-16.0)& 0.3  \\
        & {Naive-LLM} & 15.7 (16.3) & 25.5 (22.6)  &  449.2 (5.2) & 2111.9 (-11.7)& 0.8  \\
        & {GPL-LLM} & 15.9 (17.8)  & 25.9 (24.5) & 450.3 (5.0) & 2115.3 (-11.9)  & 0.9 \\
        & {CoT-LLM} & 16.9 (25.2)  & 26.4 (26.9)  & 440.1 (7.1) &  2065.6 (-9.3) & 1.2   \\
		& {IRoT-LLM} & \textbf{18.1 (34.1)} & \textbf{27.3 (31.3)}  & \textbf{431.2 (9.0)} & \textbf{2010.2 (-6.3)} & \textbf{1.7}  \\
		\hline\hline

  	\multirow{8}{*}{\makecell{LLM \\ based}} & No ad hoc  &  21.1 (-)  &  27.2 (-)  &  453.2 (-) & 1813.3 (-) & -  \\ 
        \cline{2-7} 
		
		& {Random}  &  21.1 (0)  & 27.2 (0)  &  453.2 (0) & 2186.1 (-20.1) & 0  \\
		& {Heuristic} & 23.1 (9.4) & 33.1 (21.7) &  417.6 (7.9) & 2013.3 (-11.0) & 1.8  \\
		& {PLASTIC} &  21.5 (1.9) & 28.0 (2.9) & 440.2 (2.9) & 2130.6 (-17.5) & 0.6  \\
	  & {Naive-LLM} & 23.3 (10.4)  & 33.5 (23.2)  &  418.1 (7.7) & 2015.2 (-11.1) & 1.8  \\	
        & {GPL-LLM)} & 23.5 (11.4) & 33.7 (23.9) & 415.5 (8.3)  & 2013.5 (-11.0) & 1.9 \\
        
        & {CoT-LLM} & 24.9 (18.0)  & 33.7 (23.9)  & 406.6 (10.3) & 1945.3 (-7.3) & 2.5  \\
		& {IRoT-LLM} & \textbf{26.3 (24.6)} & \textbf{35.1 (29.0)}  & \textbf{399.3 (11.9)} & \textbf{1918.1 (-5.8)} & \textbf{3.2}  \\

        \Xhline{0.6pt}
  
	\end{tabular}
	\vspace{-0.8em}
\end{table}

\renewcommand{\arraystretch}{0.8}
\begin{table}[!t] 
	\setlength{\abovecaptionskip}{-0.15cm}
	\setlength{\tabcolsep}{0.60mm}
	
	\fontsize{6.8}{9.0pt}\selectfont
	\caption{Quantitative Results in Ad hoc Teamwork with Different Team Policies when \textcolor{red}{$t_0=50$}}
	\label{quanres2}
	\centering
	 
	\begin{tabular}{c|c|c|c|c|c|c}
		\Xhline{0.6pt}
  
		\multirow{1}{*}{Team} & \multirow{1}{*}{ad hoc} & \multicolumn{1}{c|}{$\%Suc(\uparrow)$} & \multicolumn{1}{c|}{$\%PS(\uparrow)$}& \multicolumn{1}{c|}{$\#TS(\downarrow)$}  & \multicolumn{1}{c|}{$\#AS(\downarrow)$} & $\%SE(\uparrow)$ \\  
		
		
		\Xhline{0.3pt}
  
        \multirow{8}{*}{\makecell{Heuristic \\ based}} & No ad hoc  & 15.3 (-)  &  21.7 (-)  & 367.3 (-) & 1863.2 (-) &  -  \\ 
        \cline{2-7} 
		
		& {Random}  & 15.3 (0)  & 21.7 (0) & 467.3 (0) & 2183.2 (-17.2) & 0  \\
		& {Heuristic} & 17.4 (13.7)  & 26.7 (23.0)  & 439.2 (6.0) & 2001.3 (-7.4) & 1.1   \\
		& {PLASTIC}  & 15.5 (1.3) & 22.1 (1.8)  & 461.3 (1.3)  & 2132.9 (-14.5)  & 0.3    \\
        & {Naive-LLM}  & 17.2 (12.4) & 27.0 (24.4)  & 440.8 (5.7) & 2006.2 (-7.7) & 1.1  \\
        & {GPL-LLM}  & 17.4 (13.7) & 27.2 (25.3) & 436.1 (6.7) & 1996.3 (-7.1) & 1.1 \\
		
        & {CoT-LLM}  & 18.6 (21.6)  & 27.6 (27.2) & 423.2 (9.4) & 1936.3 (-3.9) &  1.8  \\
		& {IRoT-LLM} & \textbf{20.3 (32.7)}  &  \textbf{29.5 (35.9)}  & \textbf{415.8 (11.0)} & 1873.1 (-0.5) & 2.3   \\
        \hline \hline

        \multirow{8}{*}{\makecell{Learning \\ based}} & No ad hoc  & 13.5 (-)  &  20.8 (-)  & 473.8 (-) & 1890.2 & -   \\ 
        \cline{2-7} 
		
		& {Random}  & 13.5 (0)  &  20.8 (0)  &  437.8 (0) & 2176.2 (-15.1) & 0  \\
		& {Heuristic}  & 15.3 (13.3)  & 24.7 (18.8) & 458.2 (3.3) & 2092.1 (-10.7) & 0.6  \\
		& {PLASTIC}  &  13.8 (2.2)   & 22.3 (7.2)  &  465.1 (1.8)  & 2156.9 (-14.1) & 0.3    \\
		& {Naive-LLM}  &  15.3 (13.3)  &  25.1 (20.7) &  452.1 (4.6) & 2052.7 (-8.6) & 0.6  \\
        & {GPL-LLM}  & 15.6 (15.6) & 25.2 (21.2) & 452.1 (4.6) & 2050.9 (-8.5) & 0.7 \\
        & {CoT-LLM}  &  16.1 (19.3)  &  26.0 (25.0)  &  445.3 (6.0) & 2016.2 (-6.7) & 1.0 \\
		& {IRoT-LLM}  &  \textbf{17.7 (31.1)}  &  \textbf{26.8 (28.8)} &  \textbf{435.9 (8.0)} & \textbf{1995.3 (-5.6)} & \textbf{1.5}  \\
        \hline \hline

        \multirow{8}{*}{\makecell{LLM \\ based}} & No ad hoc  & 21.1 (-) & 27.2 (-) & 453.2 (-) & 1813.1 (-)& -   \\ 
        \cline{2-7} 
		
		& {Random}  & 21.1 (0) & 27.2 (0) & 453.2(0) & 2133.3 (-17.6) & 0 \\
		& {Heuristic}  & 22.5 (6.6) & 32.1 (18.0) &  423.3 (6.6) & 1988.6 (-9.7)& 1.5    \\
		& {PLASTIC}  &  21.3 (0.9)  & 27.5 (1.1)  & 443.7 (2.1)  & 2081.3 (-14.8) & 0.5    \\
        & {Naive-LLM}  & 22.5 (6.6)  & 32.6 (19.9)  &  425.3 (6.2) & 1986.1 (-9.5) & 1.4   \\
        & {GPL-LLM}  & 22.7 (7.6) & 32.9 (21.0) & 423.1 (6.6) & 1988.0 (-9.6)& 1.5 \\
        & {CoT-LLM}  & 24.0 (13.7)  & 33.1 (21.7)  & 412.1 (9.1)  & 1891.9 (-4.3) & 2.2   \\
		& {IRoT-LLM}  &  \textbf{25.6 (21.3)}  & \textbf{34.2 (25.7)}  &  \textbf{403.9 (10.9)}  & \textbf{1859.2 (-2.5)} & \textbf{2.6}   \\
        
		\Xhline{0.6pt}
		
	\end{tabular}
	\vspace{-0.8em}
\end{table}

\renewcommand{\arraystretch}{0.8}
\begin{table}[!t] 
	\setlength{\abovecaptionskip}{-0.15cm}
	\setlength{\tabcolsep}{0.60mm}
	
	\fontsize{6.8}{9.0pt}\selectfont
	\caption{Quantitative Results in Ad hoc Teamwork with Different Team Policies when \textcolor{red}{$t_0=100$}}
	\label{quanres3}
	\centering
	
	\begin{tabular}{c|c|c|c|c|c|c}
		\Xhline{0.6pt}
		
		\multirow{1}{*}{Team} & \multirow{1}{*}{ad hoc} & \multicolumn{1}{c|}{$\%Suc(\uparrow)$} & \multicolumn{1}{c|}{$\%PS(\uparrow)$}& \multicolumn{1}{c|}{$\#TS(\downarrow)$}  & \multicolumn{1}{c|}{$\#AS(\downarrow)$} & $\%SE(\uparrow)$ \\ 
		
  
		\Xhline{0.3pt}

  
        \multirow{8}{*}{\makecell{Heuristic \\ based}} & No ad hoc  & 15.3 (-)  & 21.7 (-)  & 467.3 (-)  & 1863.2 (-) & - \\ 
        \cline{2-7} 
		
		& {Random}  & 15.3 (0)  & 21.7 (0)  &  467.3 (0) & 2152.9 (-15.6) & 0   \\
		& {Heuristic}  & 16.6 (8.5)  &  25.9 (19.4)  &  442.1 (5.4) & 1967.3 (-5.6) & 1.0  \\
		& {PLASTIC}  &  15.4 (0.7)   & 21.9 (0.9)  & 465.2 (0.4) & 2107.2 (-13.1) & 0.2  \\
		& {Naive-LLM}  & 16.9 (10.5)  &  26.3 (21.2) &  443.3 (5.1) & 1955.9 (-5.0) &  0.9 \\
        & {GPL-LLM} & 17.1 (11.8)  & 26.3 (21.2) & 438.6 (6.1) & 1945.6 (-4.5) & 1.0 \\
        
        & {CoT-LLM}  &  18.0 (17.6)  &  27.3 (25.8) & 427.9 (8.4) & 1877.3 (-0.8) & 1.5  \\
		& {IRoT-LLM} &  \textbf{19.7 (28.8)} & \textbf{27.8 (28.1)}  & \textbf{420.1 (10.1)}  & \textbf{1822.8 (2.1)} &  \textbf{2.0} \\
        \hline \hline

        \multirow{8}{*}{\makecell{Learning \\ based}} & No ad hoc  &  13.5 (-) & 20.8 (-)  & 473.8 (-) & 1890.2 (-) & -   \\ 
        \cline{2-7} 
		
		& {Random}  & 13.5 (0) &  20.8 (0) &  473.8 (0)  & 2152.9 (-13.9) & 0  \\
		& {Heuristic} & 15.0 (11.1)  &  24.1 (15.9) &  461.2 (2.7) & 2061.2 (-9.0) & 0.4 \\
		& {PLASTIC}  & 13.7 (1.5)  &  22.1 (6.3) & 469.9 (0.8) & 2133.8 (-12.9) & 0.1  \\
		& {Naive-LLM} & 15.1 (11.9)  &  24.8 (19.2) &  455.3 (3.9) & 2003.8 (-6.0)& 0.6  \\
        & {GPL-LLM} & 15.3 (13.3) & 24.7 (18.8) & 453.1 (4.4) & 1995.6 (-5.6) & 0.6 \\  
        & {CoT-LLM} & 15.7 (16.3)  &  25.5 (22.6) &  449.2 (2.5) & 1973.9 (-4.4)&  0.8 \\

        & {IRoT-LLM}  &  \textbf{17.2 (27.4)} & \textbf{26.1 (25.5)} &  \textbf{439.7 (7.2)}  & \textbf{1948.2 (-3.1)} & \textbf{1.2} \\ 
  
          \hline \hline

        \multirow{8}{*}{\makecell{LLM \\ based}} & No ad hoc  & 21.1 (-) & 27.2 (-)  & 453.2 (-) & 1813.3 (-)  & -  \\ 

        \cline{2-7} 
		
		& {Random}  & 21.1 (0) & 27.2 (0)  & 453.2 (0)  & 2106.3 (-16.2) &  0  \\
		& {Heuristic}  & 21.8 (3.3) & 31.9 (17.3) & 427.6 (5.6)  &  1932.1 (-6.6) &  1.3  \\
		& {PLASTIC} & 21.3 (0.9) &  27.3 (0.4) & 449.1 (0.9) & 2067.3 (-14.0) &  0.2  \\
		& {Naive-LLM}  & 22.0 (4.3)  & 31.8 (16.9) & 428.1 (5.5)  & 1933.9 (-6.7) & 1.3  \\
  
        & {GPL-LLM} & 22.0 (4.3)  & 31.9 (17.3) & 426.6 (5.9) & 1930.9 (-6.5)  & 1.3 \\
        
        & {CoT-LLM}  &  23.2 (10.0)  & 32.3 (18.8) &  415.3 (8.4) & 1840.2 (-1.5) & 2.0  \\
		& {IRoT-LLM}  &  \textbf{25.0 (18.5)} & \textbf{33.1 (21.7)} &  \textbf{407.5 (10.1)}  & \textbf{1803.1 (0.6)} & \textbf{2.6} \\
        
		\Xhline{0.6pt}
		
	\end{tabular}
	\vspace{-0.8em}
\end{table}

\textit{Heterogeneous teammates generation}: The characteristics of the teammate are described by a tuple of $(\alpha_{nav}, \alpha_{manip}, h, m, step)$ where $\alpha_{nav}$ and $\alpha_{manip}$ indicate the navigation ability and manipulation ability of the agent, $h$ is the height of the agent, $m$ denotes the payload which means the maximum weight of the object that can be picked up by the agent, and $step$ is the max steps that the agent's battery can support. For each tidying-up scenario, we consider situations of having 3, 4, and 5 teammates, respectively, and we randomly generate the values for each teammate's characteristics. It is guaranteed that all generated agents have the navigation ability and at least one agent has manipulation ability in order to accomplish the tidying up task. Additionally, the initial positions of each teammate are randomly generated from the reachable positions in the corresponding task. 

\textit{Ad hoc agent generation}: Similar to the process of generating heterogeneous teammates. We generate one ad hoc agent for each tidying-up task. It is noted that the ad hoc agent does not have to have the manipulation ability. Its initial position is also randomly generated.

Finally, we obtain 1350 tidying-up tasks for the heterogeneous ad hoc teamwork. We divide these tasks into three difficulty categories: \textit{Easy}, \textit{Medium} and \textit{Difficult}, in which houses with 1-3 rooms are \textit{Easy} tasks, houses with 4-6 rooms are \textit{Medium} tasks, and houses with 7-10 rooms are \textit{Difficult} tasks. The heterogeneous agent and scene settings are shown in Fig. \ref{fig:dataset}. The process of generating the data can be applied and extended to the whole testing set in ProcTHOR-10K. More detailed information can be found in Appendix.

It is worth noting that the tidying-up task is different from the rearrangement task. In the rearrangement task, the agent only needs to re-place objects according to the given initial state, while in the tidying-up task, the agent is required to infer the reasonable locations of each object by itself without extra guidance and re-place misplaced objects to reasonable locations, which puts forward higher requirements for scene understanding and commonsense reasoning.

\subsection{Heterogeneous Teammates Policy}
Before the ad hoc agent joins the team, the original heterogeneous teammates collaborate with each other with a certain policy. The proposed ad hoc teamwork framework can adapt to different heterogeneous teammates no matter what policy they employ. In the experiment, we consider the following different common policies for the original heterogeneous teammates' collaboration.

\subsubsection{\textbf{Heuristic-based Policy}}
A heuristic-based policy is designed for the tidying-up task. When a teammate with manipulation ability detects the misplaced object, it will directly choose to pick up the detected misplaced object and place it in a reasonable location. Teammates without manipulation ability will continue exploring the scene. The details of the heuristic-based policy are illustrated in the Appendix.

\subsubsection{\textbf{Learning-Based Policy}}
We utilize imitation learning to train the learning-based teammate's policy to predict the sub-skill for the teammate to execute. The details of this method are shown in the Appendix.

\subsubsection{\textbf{LLM-based Policy}}
An LLM-based policy is designed for heterogeneous teammates to collaborate. The teammate utilizes the LLM to generate the next sub-task intention and negotiate with other teammates to reach an agreement according to the current state. Then, the LLM predicts the sub-skill for the agent to execute. The details of the LLM-based policy are illustrated in Appendix.

\begin{figure*}[htbp]
	\centering
	\subfigure[Performance in $\%Suc$ of different ad hoc policies.]
    {\label{a}
    \includegraphics[width=2.3in, height = 1.5in]{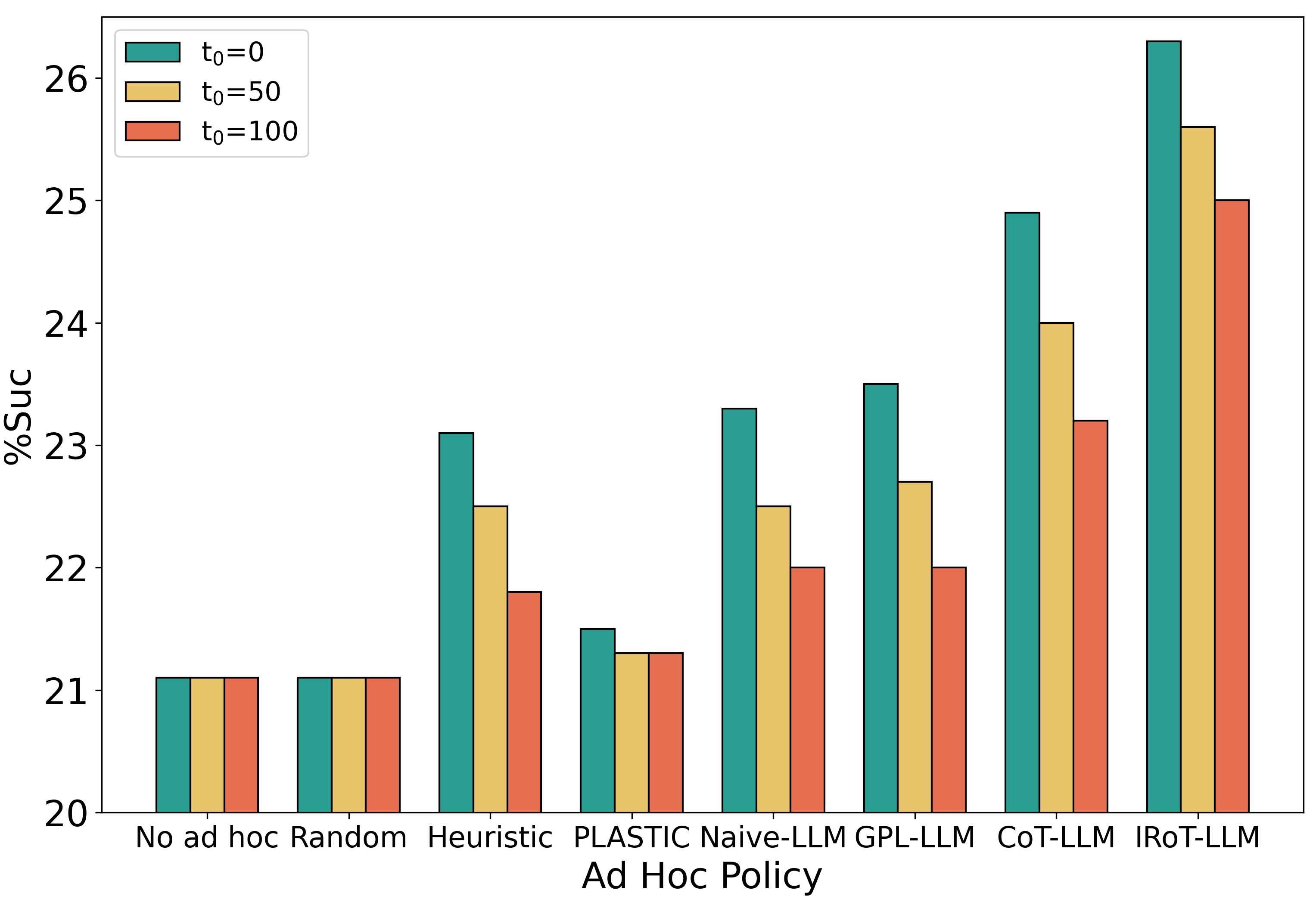}
    }
	\subfigure[Performance in different task categories.] 
     {\label{b}
     \includegraphics[width=2.25in, height = 1.5in]{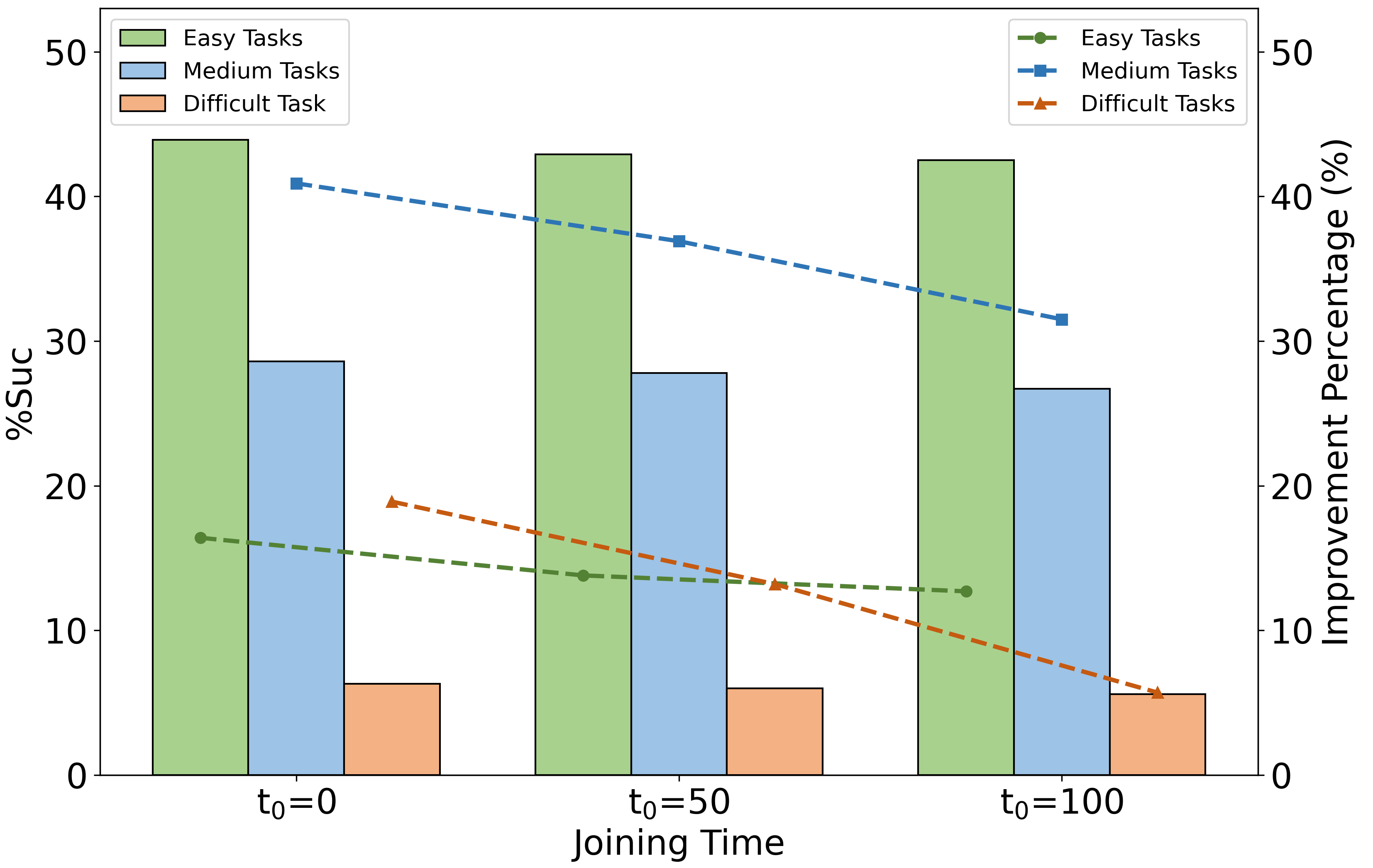}
     }
     \subfigure[Performance in different ad hoc ability.] 
     {\label{c}
     \includegraphics[width=2.25in, height = 1.5in]{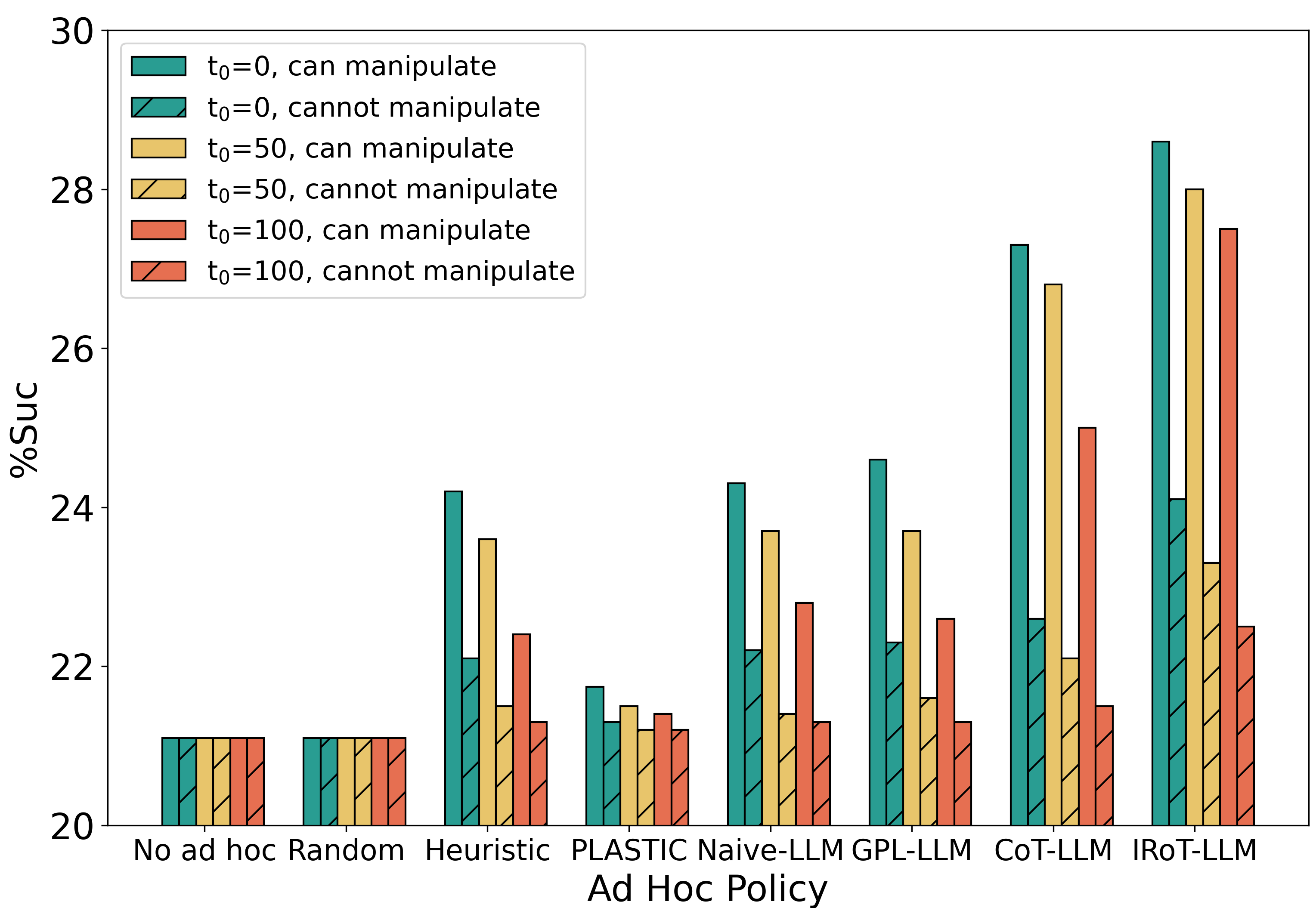}}
	\caption{The performance of different ad hoc policies. Fig. \ref{a} shows the performance of different ad hoc policies collaborating with the LLM-based teammates when joining at different times. Fig. \ref{b} demonstrates the performance of our IRoT-LLM policy collaborating with the LLM-based teammates in different categories of tasks. The histogram represents the value of $\%Suc$, and the points on the line chart represent the improvement percentage compared with teammates with no ad hoc agent. Fig. \ref{c} shows the performance of the ad hoc policies when the ad hoc agent has different abilities.}
	\label{quan_fig}
\end{figure*}

\subsection{Ad hoc Agent Policy}
For the ad hoc agent, we compare the proposed method with several baselines as follows.

\subsubsection{\textbf{Random Policy}}
The ad hoc agent randomly chooses one action from its action space to execute at each step.

\subsubsection{\textbf{Heuristic-based Policy}}
Similar to the heuristic-based for the original heterogeneous teammates, for the ad hoc agent with manipulation ability, if a misplaced object is detected, it will directly pick it up and place it in a reasonable location. For the ad hoc agent without manipulation ability, it will directly explore the scene from its current location. 

\subsubsection{\textbf{PLASTIC Policy}}
We train the ad hoc policy with several pre-defined teams based on the existing common method of \textit{PLASTIC} \cite{barrett2017making} and the predicted sub-skill is executed with imitation learning. The ad hoc agent directly uses the learned policy to collaborate with unknown heterogeneous teammates. 

\subsubsection{\textbf{Naive-LLM Policy}}
We utilize the LLM to directly predict the sub-task for the ad hoc agent without specific prompt engineering for thinking and reasoning. The sub-skill planning and execution is the same as the proposed policy.

\subsubsection{\textbf{GPL-LLM Policy}}
Most of the existing learning methods for ad hoc teamwork are suitable for a simple 2D environment, and almost no existing method can be directly applied to a 3D environment because of the substantially increased dimensions of visual information and the complex characteristics of partial observation and interaction with the environment. The latest Graph-based Policy Learning method \cite{rahman2021towards} is also difficult to be directly transferred to our task, but we also hope to verify the effect of this idea in our ad hoc task, so we combine the idea with the LLM to form the baseline \textit{GPL-LLM}. We utilize the LLM to first predict the next sub-task and sub-skill of all teammates based on the current states and capabilities of all agents. We let the LLM generate candidate sub-tasks that can be executed by the ad hoc agent and give the evaluation score for each generated sub-task according to the predicted plans of teammates. The final sub-task for the ad hoc agent to execute is generated according to the evaluation score.

\subsubsection{\textbf{CoT-LLM Policy}}
We utilize the CoT prompts to predict sub-tasks for the ad hoc agent without the evaluation and communication reflection mechanism. The sub-skill planning and execution are the same as our proposed policy.

\subsection{{Implementation Details}}

{We call our method based on IRoT as the \textbf{\textit{IRoT-LLM}} model. All the LLM-based baselines and our \textit{IRoT-LLM} model utilize the \textit{gpt-4-1106-preview} to complete the planning process. The parameter \textit{Temperature} is set to be 0.3 in our experiments. The maximum limited number of tokens in the prompt is 128,000. If the maximum limit is exceeded in tasks, we would move out some of the earliest information from the historical communication messages in the prompt to adapt to the maximum limit.}

\subsection{Evaluation Metrics}
\label{EM}
The effectiveness of an ad hoc policy can be quantified by the difference in performance between the original heterogeneous team and the new team after the ad hoc agent joins. We denote $N_{epi}$ as the number of tasks. In the context of our experiment, we utilize the evaluation metrics as follows:

\textit{1) \%Suc}: The success rate of tasks. $\%Suc=\frac{1}{N_{epi}}\sum_{i=1}^{N_{epi}}R_i$, where $R_i=1$ if all misplaced objects are re-placed in reasonable locations, otherwise $R_i=0$. It denotes the strict success rate.

\textit{2) \%PS}: The partial success rate of tasks. $\%PS=\frac{1}{N_{epi}}\sum_{i=1}^{N_{epi}}\frac{K^{suc}_i}{K_i}$, where ${K_i}$ is the number of misplaced objects in the $i$-th task, and $K^{suc}_i$ is the number of misplaced objects that are re-placed to reasonable locations. It reflects the proportion of misplaced objects that are successfully tidied.

\textit{3) \#TS}: The average temporal steps of task completion. $\#TS=\frac{1}{N_{epi}}\sum_{i=1}^{N_{epi}}TS_i$, where $TS_i$ is the number of temporal steps when completing the $i$-th task, the value equals the maximum steps among all agents' steps when they stop. If agents do not successfully complete the task, $TS_i$ is the maximum number of steps. It evaluates the temporal efficiency of completing the task. The fewer $\#TS$, the higher the efficiency.

\textit{4) \#AS}: The total action steps of all agents in tasks. $\#AS=\frac{1}{N_{epi}}\sum_{i=1}^{N_{epi}}AS_i$, where $AS_i$ is the sum of action steps of all agents when they stop in the $i$-th task. If agents do not successfully complete the task, the action step of each agent is the maximum number of steps. It can evaluate the total energy consumption after the ad hoc agent joins the team. The fewer $\#AS$, the less energy consumption of the collaboration.

\textit{5) \%SE}: Success rate weighted by efficiency improvement. $SE=\frac{1}{N_{epi}}\sum_{i=1}^{N_{epi}}\frac{max(0, TS_i^{team}-TS_i^{adhoc})}{TS_i^{team}}{R_i^{adhoc}}$, where $TS_i^{team}$ is the number of temporal steps when only teammates completing the $i$-th task, and $TS_i^{adhoc}$ is the number of temporal steps when the ad hoc agent joins the team to complete the $i$-th task. It can evaluate the efficiency improvement of the ad hoc agent collaboration compared with the existing team.

\begin{figure*}
    \centering  
	\includegraphics[width=7in]{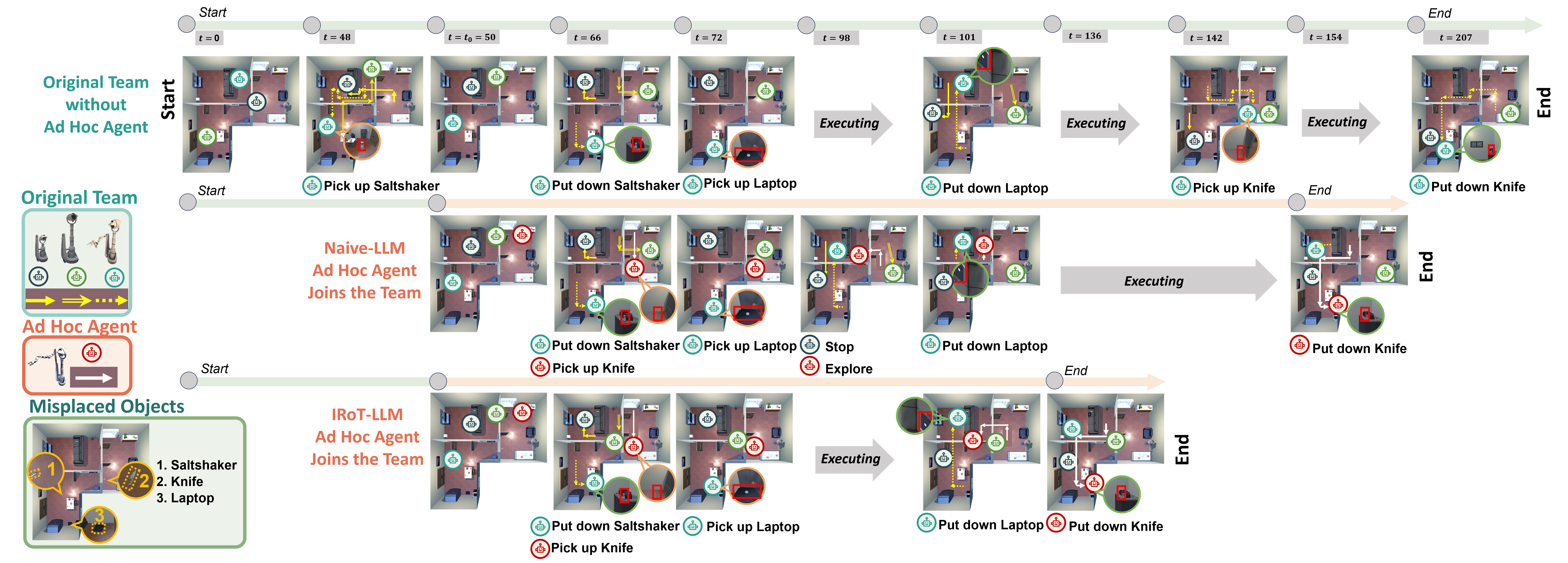}
	\caption{The sample of the ad hoc agent collaborating with the LLM-based original team. The icons of different colors mark positions of agents. Yellow lines with different shapes mark the paths between key steps of teammates, and white lines mark the paths of the ad hoc agent. {A main timeline from 0 to 207-th step is illustrated at the top. The first row represents the task execution process of the original team in the top-down view. The second row shows the execution process of a new team with the \textit{Naive-LLM} ad hoc agent joining at the 50-th step. The third row outlines the process of another new team with the \textit{IRoT-LLM} ad hoc agent also joining at the 50-th step.} The original team tidies up these three misplaced objects at the 207-th step. The new team with the \textit{Naive-LLM} ad hoc agent finishes this task at the 154-th step, while the new team with the \textit{IRoT-LLM} ad hoc agent finishes this task at the 136-th step. \textit{IRoT-LLM} performs better than \textit{Naive-LLM}, since \textit{Naive-LLM} predicts the next sub-task without evaluation and reflection when re-planning. \textit{Naive-LLM} switches to \textit{Explore} at the 90-th step when re-planning after a teammate stops and then switches back to the sub-task \textit{Re-place} after a few steps, while the \textit{IRoT-LLM} one continuing executing the sub-task \textit{Re-place} until it puts down the object, which demonstrates the effectiveness of the proposed IRoT method.}
	\label{fig:simulation}
\end{figure*}

\subsection{Quantitative Performance}
We evaluate the performance of our ad hoc framework collaborating with different policies of teammates at different times; the results of joining the team at $t_0=0$, $t_0=50$ $t_0=100$ are illustrated in Table \ref{quanres1}, \ref{quanres2} and \ref{quanres3} respectively. We call our method based on IRoT as the \textbf{IRoT-LLM} policy. We calculate the value improvement percentage for the metrics that are better with the higher value ($\%Suc$, $\%PS$), and calculate the decline percentage for the metrics that are better with the lower value ($\#TS$, $\#AS$), and put this value in brackets. The performance shows the influence of different ad hoc methods on the original team when the ad hoc agent joins the team and collaborates with teammates, and from the results we obtain the following findings:

1) \textbf{Our framework can obviously improve the success rate and efficiency of the original team of different policies compared with no ad hoc agent joining the team.} Our method improves $\%Suc$, $\%PS$, $\%SE$ and reduces $\#TS$, $\#AS$ when collaborating with the heuristic-based, learning-based, and LLM-based teams, which demonstrates that our policy can adapt to different policies of teams. The improvements on the original team of our policy differ in collaborating with a team of different policies, and the improvement can be larger when the team policy itself is worse. The performance of different ad hoc policies in $\%Suc$ collaborating with the LLM-based team at different times is also illustrated in Fig. \ref{a}.

2) \textbf{Our framework can improve the performance of the original teams of different policies when joining them at different times} of $t_0=0$, $t_0=50$, and $t_0=100$. It is noticed in the Fig. \ref{a}, the later the ad hoc agent joins the original team, the less improvement of the ad hoc agent on this team. This is because teammates have already cooperated with each other and completed more parts of the task when the ad hoc agent joins the team later. The results demonstrate that the ad hoc agent with the proposed policy can still adapt to the team no matter when it joins the team and improve the performance of the team in the subsequent execution of the task.

\renewcommand{\arraystretch}{0.8}
\begin{table}[!t] 
	\setlength{\abovecaptionskip}{-0.15cm}
	\setlength{\tabcolsep}{0.60mm}
	
	\fontsize{6.8}{9.0pt}\selectfont
	\caption{Results in Ablation Experiments of Ad hoc Teamwork}
	\label{ablation}
	\centering
	
	\begin{tabular}{c|c|c|c|c|c|c}
		\Xhline{0.6pt}
		
		\multirow{1}{*}{$t_0$}  & \multirow{1}{*}{Ad hoc Policy} & \multicolumn{1}{c|}{$\%Suc(\uparrow)$} & \multicolumn{1}{c|}{$\%PS(\uparrow)$}& \multicolumn{1}{c|}{$\#TS(\downarrow)$}  & \multicolumn{1}{c|}{$\#AS(\downarrow)$} &$\%SE(\uparrow)$ \\ 
		

  
		\Xhline{0.5pt}
		\multirow{6}{*}{0} & No ad hoc  &  21.1 (-)  &  27.2 (-)  & 453.2 (-)  &  1813.3 (-)  & - \\ 
        \cline{2-7} 
		
		& {w/o STP.} & 21.9 (3.8) & 27.8 (2.2) & 441.3 (2.6) & 2183.2 (-20.4) & 0.6 \\
        & {IRoT w/o Ev. \& Rf.} & 22.5 (6.6)  & 28.3 (4.0)  &  423.2 (6.6)  & 2102.3 (-15.9)  & 1.5  \\
		& {IRoT w/o Ev.} & 23.9 (13.3)  & 31.8 (16.9) &  417.3 (7.9)   & 2013.9 (-11.1) & 1.9\\
		& {IRoT w/o Rf.}  & 23.6 (11.8)  & 31.3 (15.1)  &  418.8 (7.6)  & 2020.1 (-11.4)  &  1.8 \\
        & {IRoT (GPT-3.5)}  & 25.0 (18.5)  &  33.2 (22.1) &  403.9 (10.9)  & 1956.3 (-7.9)  &  2.7 \\
		& {IRoT-LLM} & \textbf{26.3 (24.6)}  &  \textbf{35.1 (29.0)} &  \textbf{399.3 (11.9)}  & \textbf{1918.1 (-5.8)}  & \textbf{3.2}  \\
		
        \hline \hline
  
        \multirow{6}{*}{50} & No ad hoc  & 21.1 (-)  &  27.2 (-)  & 435.2 (-)  & 1813.3 (-)  & - \\ 
        \cline{2-7} 
		
		& {w/o STP.}  & 21.6 (2.4)   & 27.7 (1.8)  &  446.1 (1.6)  & 2132.8 (-17.6) & 0.4  \\
        & {IRoT w/o Ev. \& Rf.}  & 22.3 (5.7)  & 28.0 (3.0)  &  441.3 (2.6)  & 2050.6 (-13.1) &  0.6 \\
		& {IRoT w/o Ev.}  & 23.5 (11.4)   & 31.3 (15.1)  &  425.2 (6.2) & 1965.7 (-8.4)  & 1.4  \\
		& {IRoT w/o Rf.}  & 23.4 (10.9)  & 31.0 (14.0) &  427.1 (5.8)  & 1970.7 (-8.7)  & 1.3  \\
        & {IRoT (GPT-3.5)}  & 24.8 (17.5)  & 33.0 (21.3)  &  408.2 (9.9)  & 1903.1 (-5.0)  & 2.5  \\
		& {IRoT-LLM}  & \textbf{25.6 (21.3)}  &  \textbf{34.2 (25.7)}  &  \textbf{403.9 (10.9)} & \textbf{1859.2 (-2.5)} & \textbf{2.6}  \\
        \hline \hline

        \multirow{6}{*}{100} & No ad hoc & 21.1 (-)  &  27.2 (-)  & 435.2 (-)  & 1813.3 (-)  & -  \\ 
        \cline{2-7} 
		
		& {w/o STP.}  & 21.5 (1.9)   & 27.5 (1.1)  &  450.8 (0.5)  & 2103.5 (-16.0) &  0.2 \\
        & {IRoT w/o Ev. \& Rf.}  & 22.0 (4.3)  & 27.7 (1.8) &  446.6 (1.5)  & 2006.1 (-10.6)  & 0.3  \\
		& {IRoT w/o Ev.}  & 22.9 (8.5)   & 20.8 (13.2)  &  429.8 (5.2)  & 1911.3 (-5.4)  & 1.2  \\
		& {IRoT w/o Rf.}  & 22.7 (7.6)  & 20.6 (12.5)  &  431.2 (4.9)  &  1920.9 (-5.9)  & 1.1  \\
        & {IRoT (GPT-3.5)}  & 23.8 (12.8)  & 31.8 (16.9)  &  412.3 (9.0)  & 1856.2 (-2.4)  &  2.2 \\
		& {IRoT-LLM}  & \textbf{25.0 (18.5)}  &  \textbf{33.1 (21.7)} & \textbf{407.6 (10.0)} & \textbf{1803.1 (0.6)} & \textbf{2.6}  \\

		\Xhline{0.6pt}
	\end{tabular}
	\vspace{-1.2em}
\end{table}

3) \textbf{Our framework performs better than other ad hoc policies when collaborating with different teams.} The ad hoc policies based on the LLM including \textit{Naive-LLM}, \textit{GPL-LLM}, \textit{CoT-LLM} and \textit{IRoT-LLM} achieve higher $\%Suc$, $\%PS$ and $\%SE$ as well as lower $\#TS$, $\#AS$ than the learning-based baseline \textit{PLASTIC}, since LLM-based methods can deal with unseen complex circumstance more flexible, and have better generalization ability. \textit{Naive-LLM} performs similarly to \textit{Heuristic-based} policy, which to some extent shows that planning in the embodied environment is more difficult compared with a simple static planning problem. \textit{IRoT-LLM} performs the best among all LLM-based ad hoc policies, demonstrating the effectiveness of the IRoT method with the communication messages. The performance of \textit{GPL-LLM} shows that although most of the existing learning methods for ad hoc teamwork are difficult to be directly transferred to our 3D task, combining the key idea with the LLM can still achieve moderate results. When the ad hoc agent joins the original team, it would improve the efficiency of the team but also might bring more energy consumption due to the addition of one more agent. The $\#AS$ of \textit{IRoT-LLM} when joining the team at $t_0=100$ is even smaller than the original team, which demonstrates the proposed method can effectively improve the efficiency of the existing team.

More specifically, we separately compare the results of the proposed policy in three difficulty categories when collaborating with the LLM-based teammates, which is shown in Fig. \ref{b}. The comparison performance illustrates that the proposed model achieved better improvement after the joining of the ad hoc agent in the \textit{Medium} tasks since the moderate number of rooms provides a good chance for the newly joined ad hoc agent to effectively help the original team and avoid conflicts with the original team cooperation policies. In \textit{Easy} tasks, the existing team can already achieve better performance, so the performance improvement is not obvious. In \textit{Difficult} tasks, the large number of rooms makes it more difficult to find and tidy all misplaced objects, so the improvement of ad hoc teamwork is also limited. Meanwhile, we further compare the influence on the existing LLM-based team and whether the joined IRoT-based ad hoc agent has the manipulation ability, which is illustrated in Fig. \ref{c}. The proposed framework can adapt to ad hoc agents with different abilities, and the results demonstrate that the ad hoc agent with the manipulation ability can better help the original teams complete the task, obviously improving efficiency.

\begin{figure}
	\centering
     \subfigure[Robots] {\label{fig:Robots}\includegraphics[width=1.66in, height=1.66in]{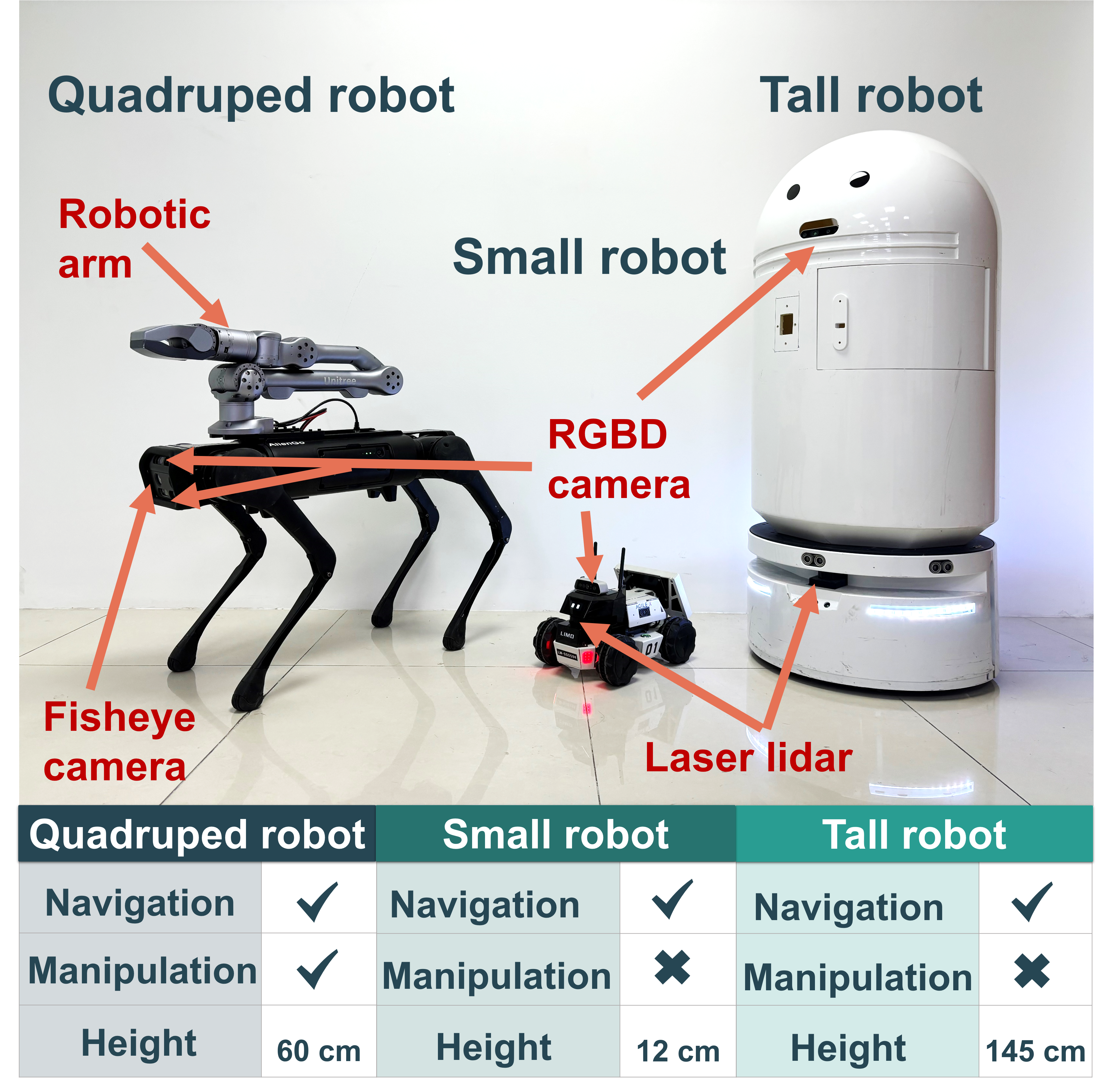}}
	\subfigure[Environments] {\label{fig:Rooms}\includegraphics[width=1.66in, height=1.66in]{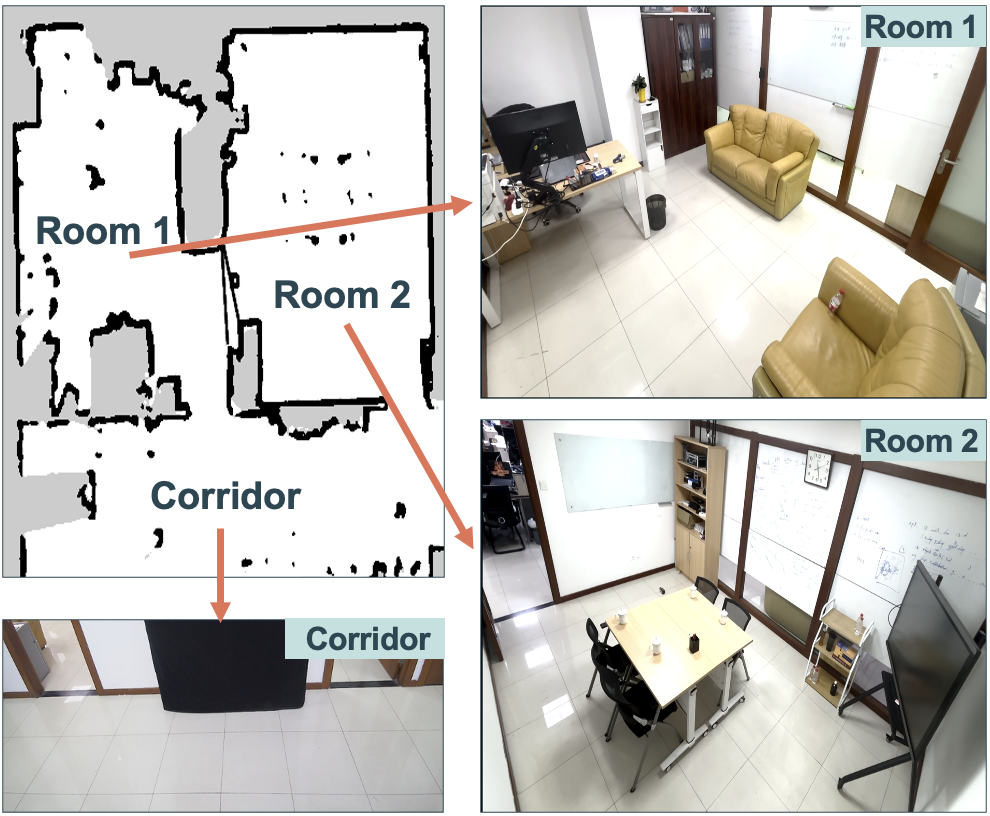}}
	\caption{The robots and environment settings of physical validation.}
   \vspace{-1.0em}
\end{figure}

\begin{figure*}
	\centering
	\includegraphics[width=7in]{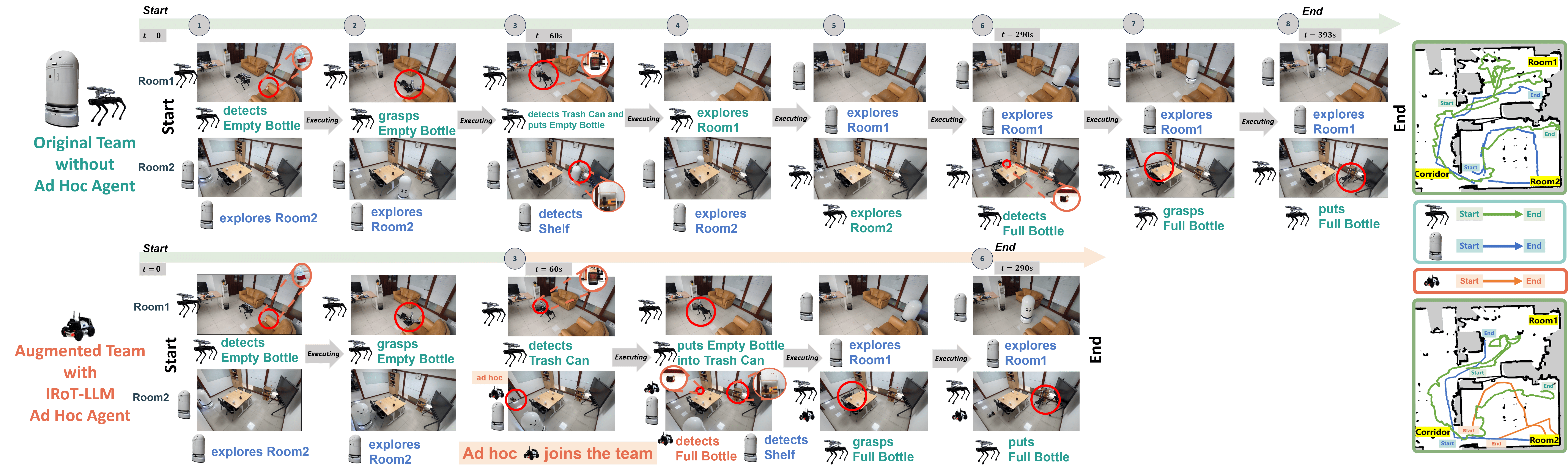}
	\caption{A successful example of the ad hoc robot collaborating with the LLM-based original team in the real-world experiment. A main timeline highlighting significant time steps from 0 to 393 seconds is shown at the top. Two rows of images demonstrate the third-person view of \textit{Room1} and \textit{Room2} and the explanation of the robots' actions at the key time step. The first row of images shows the original team tidying the rooms, while the second row shows the new team performing the same task in a much shorter time. The far right of this picture provides two top-down views of the agents’ paths while performing the task.}
	\label{fig:Flowchart}
\end{figure*}

\begin{figure*}
	\centering
	\includegraphics[width=7in]{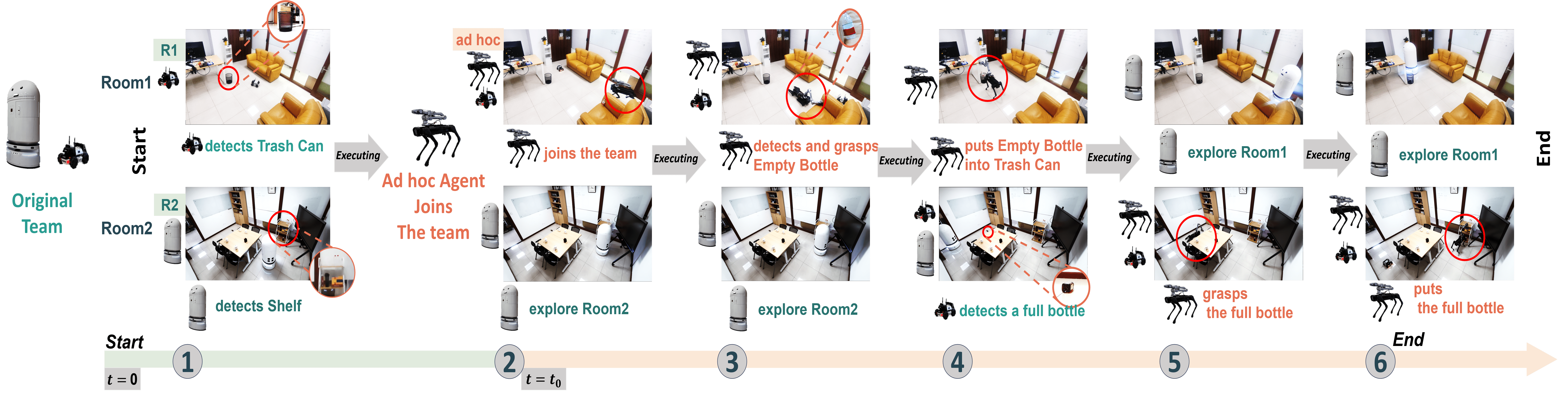}
	\caption{A successful example of the extended setting in the real-world experiment. The quadruped robot acts as the ad hoc robot and joins the team after the original team starts the task for a while. In this task, the original team cannot complete the task, and the joining of the quadruped robot helps the original team complete this task.}
	\label{fig:Flowchart2}
\end{figure*}

\subsection{Qualitative Performance}
\label{qual_per}
We show the sample of the ad hoc agent collaborating with the LLM-based original team and compare the collaboration performance of \textit{IRoT-LLM} ad hoc policy with \textit{Naive-LLM} policy, which is demonstrated in Fig. \ref{fig:simulation}. The original team contains three heterogeneous agents and completes the task with three misplaced objects at the 207-th step. The ad hoc agent joins the team at the 50-th step, and the actions of teammates are the same as the original team before the ad hoc agent joins the team. The new team with the ad hoc agent completes the task earlier than the original team. The new team with \textit{IRoT-LLM} ad hoc agent finishes the same task earlier than the \textit{Naive-LLM} policy, since \textit{Naive-LLM} generates the executable sub-task without evaluation and reflection when re-planning, which demonstrates the effectiveness of the IRoT method. We also show the failure samples of the dynamic sub-task planner of the ad hoc agent in the Appendix.

\subsection{Ablation Experiments}
We conduct ablation experiments to evaluate the effectiveness of the proposed hierarchical dynamic planner and IRoT-based sub-task planner. The comparison performance of joining the team at different times and collaborating with LLM-based teammates are shown in Table \ref{ablation}. The method \textit{w/o STP.} removes the dynamic sub-task planner and directly generates the sub-task with the LLM. The performance has decreased obviously compared with our framework, which demonstrates that hierarchical planning is effective in decision-making and can avoid invalid planning. We also discuss the effectiveness of different components of the proposed IRoT methods in the sub-task planning. \textit{IRoT w/o Ev. \& Rf.} does not utilize the evaluation \& ranking as well as the interactive reflection steps, and it directly generates $n_{irot}$ candidate sub-task and then chooses to execute them in the order of generation. \textit{IRoT w/o Ev.} and \textit{IRoT w/o Rf.} remove the evaluation and reflection steps respectively. The comparison results show that the three steps of the proposed IRoT method are all useful, and each step can provide effective information for sub-task planning. \textit{IRoT(GPT-3.5)} utilizes the \textit{gpt-3.5-turbo-1106} model to complete the planning, and the performance demonstrates that the reasonable planning method is crucial for the LLM to fulfill its remarkable potential in planning and decision-making. The results of the ablation experiments show the effectiveness and efficiency of the proposed hierarchical planning with the IRoT planner.

\section{Physical validation}

Due to the remarkable generalization ability of the LLM with the proposed IRoT method, the proposed ad hoc policy can be applied to the real-world scenario. To validate the effectiveness and generalization ability of our method in real-world scenarios, we utilize three heterogeneous robots for the experimental procedure in the tidying-up task, as illustrated in Figure \ref{fig:Robots}. It’s crucial to highlight that these three agents demonstrate considerable differences in their visual range and manipulation capacities. The agent on the far left is a quadruped robot with a manipulator, followed by a small four-wheeled robot in the center, and the last agent is a tall robot engineered for indoor navigation.The experimental environment is configured as two rooms connected by a corridor, as shown in Fig. \ref{fig:Rooms}. In the real-world experiments, each robot has a computer on itself to process the information from the RGB-D camera and the Laser Radar, record its movement path and control the robot to execute the corresponding actions. There also exists another computer to cache the communication messages and act as the intermediate node to exchange communication information among robots. Each robot records its trajectory in real time. The average latency for receiving responses from LLMs is approximately two seconds when planning the sub-task and the sub-skill. Unlike the simulation experiment, where agents synchronously execute actions at each timestep, in real-world experiments, robots execute actions asynchronously. It allows robots to continue executing actions while others are waiting for the LLM response.

We establish an original team consisting of the quadruped robot and the tall robot, introducing the small robot into the environment after the task starts for a period of time. The original team utilizes the \textit{LLM-based Policy} as the collaboration strategy, and the ad hoc robot utilizes the \textit{IRoT-LLM Policy}. As shown in Fig. \ref{fig:Flowchart}, during this task, the quadruped robot identifies and picks up an empty bottle from the sofa and then puts it in the trash can. When no ad hoc robot joins the original team, the quadruped robot needs to search and locate the misplaced full bottle itself after it completely explores \textit{Room1}, because the tall robot cannot see the full bottle on the floor. However, when the small robot joins the original team as an ad hoc robot after 60 seconds, it can detect the full bottle sooner and send this message to the quadruped robot, which helps the quadruped robot find the full bottle and put it on the shelf faster. The whole task finishes once the quadruped robot puts the full bottle back to the shelf. The execution process of this task demonstrates that with the assistance of the ad hoc robot, the efficiency of the task is obviously increased compared with the performance of the original team. The paths of the original teammates are also decreased compared with the original paths.

{We also consider another extended setting in which the original team cannot complete the task in the real-world scenario in Fig. \ref{fig:Flowchart2}. In this setting, the original team consists of the tall robot and the small robot, and the quadruped robot serves as the ad hoc robot. After the original team starts the task for a period of time, the quadruped robot joins the team to collaborate with other robots. Since all robots in the original team do not have the manipulation ability, they cannot complete this task. After the quadruped robot joins the team, the new team can complete this task with information from the original team. The quadruped robot picks up the empty bottle and puts it to the trash can, then it enters the \textit{Room2} to picks up the full bottle with the communication information from the small robot and puts it on the shelf. This sample demonstrates that the proposed ad hoc framework can not only solve the ad hoc teamwork task in the case that the original team can complete the task, but also can handle the case that the original team does not have the ability to complete the task. More details and the execution process of the real-world experiments can be seen in the supplementary video.}

\section{Conclusion} 
\label{sec:conclusion}

In this paper, we focus on heterogeneous ad hoc teamwork collaboration, which mainly considers generating the reasonable policy of an ad hoc robot to collaborate with unknown teammates without prior coordination. We extend the ad hoc teamwork collaboration problem to a more general and challenging setting where the ad hoc robot can seamlessly join any team at any time from any location. We leverage the strong reasoning and generalization abilities of the LLM and propose the Interactive Reflection of Thought method to dynamically plan the sub-task and sub-skill for the ad hoc robot to execute. We also build a testing benchmark to evaluate the proposed ad hoc teamwork framework in the heterogeneous multi-agent tidying-up task. Extensive comparison and ablation experiments are conducted in the benchmark to demonstrate the effectiveness of the proposed framework. We have also employed the proposed framework on physical robots in a real-world scenario. In future work, we would like to extend this work to more general scenes, such as open-ended scenes, as well as more practical settings, such as the partial communication between robots. {We would also like to consider introducing and supporting humans as the existing teammates in the ad hoc teamwork.}

\bibliographystyle{plainnat}
\bibliography{references}

\newpage
\appendix
\renewcommand\thefigure{\Alph{section}\arabic{figure}}
\setcounter{table}{0}
\setcounter{figure}{0}

\subsection{Details of the Dynamic Sub-Task Planner}
\begin{figure}
	\centering
	\includegraphics[width=3.45in]{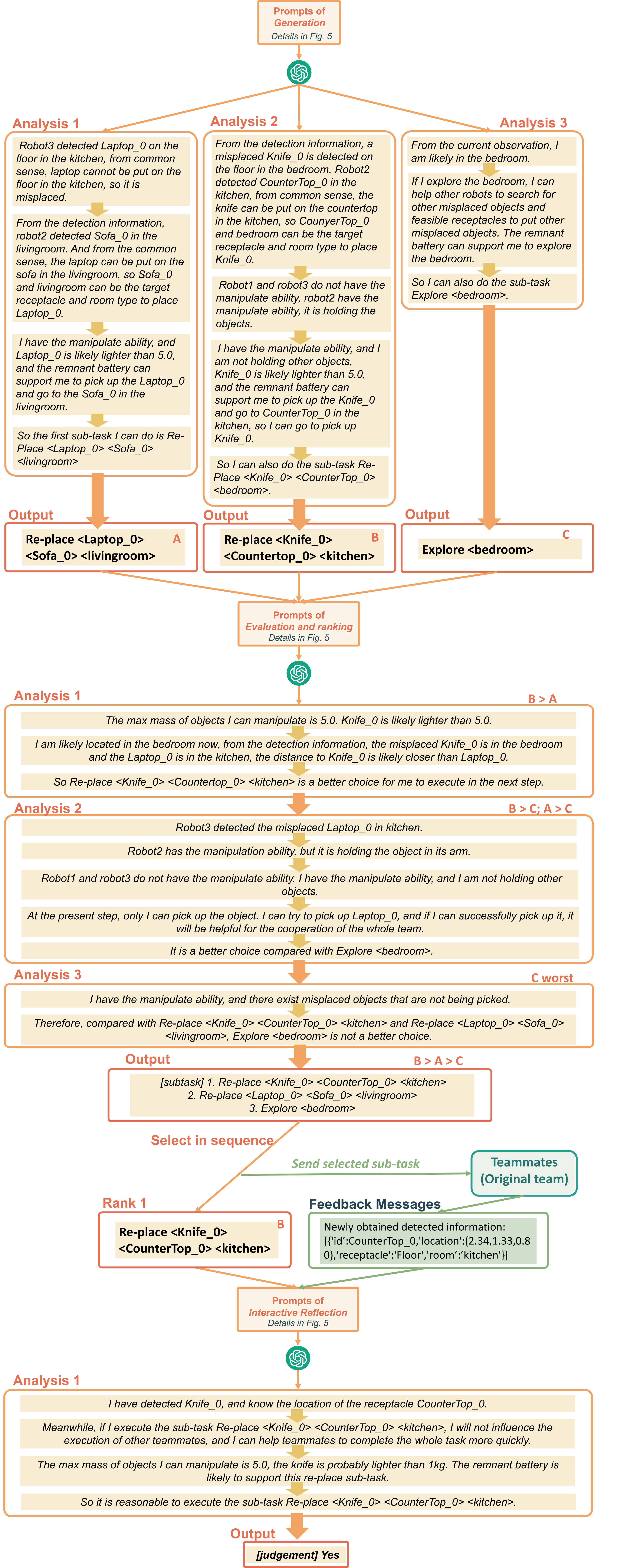}
	\caption{One example of analysis output of the dynamic sub-task planner with IRoT.}
	\label{fig:SubtaskAnalysis}
\end{figure}

We show the output example of the dynamic sub-task planner with IRoT in our proposed framework in Fig. \ref{fig:SubtaskAnalysis}. The dynamic sub-task planner consists of three steps, and the detailed analysis for each step of a sample is illustrated in Fig. \ref{fig:SubtaskAnalysis}. The first step, generation of candidate sub-tasks, outputs three sub-tasks that can be performed and gives the reasoning analysis for the output. The second step, evaluation and ranking, ranks generated three sub-tasks from good to bad and give analysis of the comparison results among these sub-tasks. The third step, interactive reflection, re-judges each sub-task from the ranked list in sequence whether the chosen sub-task is feasible to be executed considering the feedback messages from teammates.

\subsection{Details of the Dynamic Sub-Skill Planner}
\subsubsection{Description of Sub-Skills}
\label{des_subskill}

The details of candidate sub-skills that are generated by the dynamic sub-skill planner are denoted as follows.
\begin{itemize}
	\item GoToObject $\left\langle ID_{object} \right\rangle$: Navigate to the vicinity of the detected object $objectID$ in the current view.
	\item GoToPoint $\left\langle \Delta x,\Delta y \right\rangle$: Navigate to the relative position $(\Delta x,\Delta y)$ from the current position, and $\Delta x$, $\Delta y$ represent the distance to the target point in the x and y axis respectively within a certain distance constraint. This sub-skill corresponds to the embodied PointNav task.
	\item GoToRoom $\left\langle ID_{room} \right\rangle$: Navigate to the location of the known room $ID_{room}$.
	\item PickupObject $\left\langle ID_{object} \right\rangle$: Pick up the detected object $ID_{object}$ when the agent is located near it.
	\item PutObject $\left\langle ID_{objectID} \right\rangle$ $\left\langle ID_{receptacle}\right\rangle$ $\left\langle ID_{room}\right\rangle$: Place the object $\left\langle ID_{objectID} \right\rangle$ being picked up to the target receptacle $\left\langle ID_{receptacle} \right\rangle$ in the room $\left\langle ID_{room}\right\rangle$. The execution of this sub-skill requires that the agent has picked up the corresponding object.
	\item Explore: Explore the specific area to obtain more information about the scene.
	\item Stop: Stop at the current location.
\end{itemize}

\subsubsection{Output Example of the dynamic Sub-Skill planner}

\label{exp_subskill}
\begin{figure}
	\centering
	\includegraphics[width=3.45in]{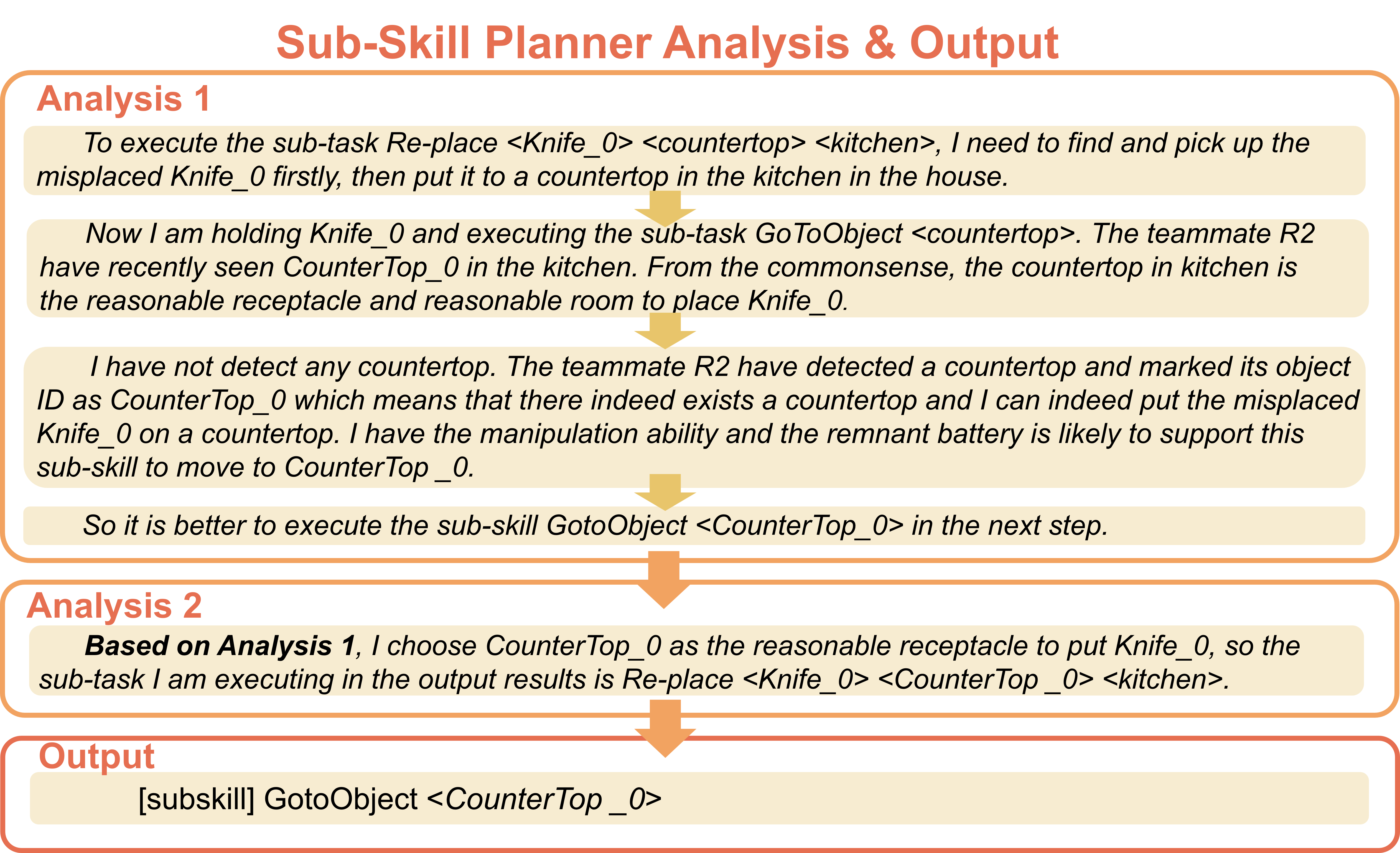}
	\caption{Example of the analysis and output of the dynamic sub-skill planner.}
	\label{fig:subskillCoT}
\end{figure}

We show an output example of the dynamic sub-skill planner in Fig. \ref{fig:subskillCoT}, which demonstrates the analysis and reasoning process of the planner.

\subsection{Prompt Template Design}

The dynamic sub-task planner and the sub-skill planner generates the next sub-task and the sub-skill with the LLM.

The dynamic sub-task planner utilizes the proposed IRoT mechanism to plan the sub-tasks. It has three stages. In the first stage \textit{Generation of Candidate Sub-Tasks}, the model generates $n_{irot}$ sub-tasks. The prompts include the task description, the candidate sub-task description, the output format, the in-context examples, the state description of the ad hoc agent, the detection information, the abilities and states of teammates, and the communication messages, as is shown in Fig. 5. The task description represents the task situation, task setting and what needs to be down to complete this ad hoc teamwork task, which gives the overall goal of the task. The candidate sub-task description introduces the categories of sub-tasks that the planner can generate from. The output format gives the format of the planning results, which restricts the output contents and is convenient for subsequent processing. The in-context examples provide the LLM with some correct samples of the output results, which makes the LLM learn and imitate the reasoning process as well as the relationship between the reasoning and the output results when generating sub-tasks. The state description of the ad hoc agent, the detection information, the abilities and states of teammates, and the communication messages are crucial information to complete the sub-task generation process. The detection information includes the historical detection information and the current detection information, which are both important to plan the next sub-task according to the historical and current state. Since the ad hoc agent and teammates are heterogeneous, the abilities of teammates are important for the ad hoc agent to plan the next sub-task. Finally, we utilize the generation prompts to inform the LLM to generate several candidate sub-tasks based on the current state and scene information. In the second stage \textit{Evaluation \& Ranking}, the model evaluates and ranks the generated sub-tasks based on their execution rationality. The prompts in this stage include the task description, the output format, the in-context examples, the state description of the ad hoc agent, the detection information, the abilities and states of teammates, the communication messages, and the generated sub-tasks, as is shown in Fig. 5. Since the generated sub-tasks are required to be sorted in this stage, the generated sub-tasks are also included in the prompts and the contents of the rest information are the same as those included in the first stage. We utilize the evaluation prompts to notify the LLM to evaluate and rank the generated sub-tasks. In the third stage \textit{Generation of Candidate Sub-Tasks}, the model judges the feasibility of the selected sub-task with the interactive communication messages from teammates. The prompts in the third stage include the task description, the candidate sub-task description, the output format, the in-context examples, the state description of the ad hoc agent, the detection information, the abilities and states of teammates, the communication messages and the chosen sub-task, as is shown in Fig. 5. Since the model needs to judge whether the selected sub-task is reasonable with the feedback communication information from teammates in this stage, the communication messages contain the updated interactive information which includes the previously detected objects or explored room related to the selected sub-task given by teammates. We utilize the rejudging prompts to let the LLM re-judge the selected sub-task with the newly obtained communication information. The prompts of the dynamic sub-task planner are designed according to the goal that the planner needs to complete in each stage, which contains the crucial descriptions of the task settings, the key state and abilities of the ad hoc agent and teammates as well as the effective communication information. The designed prompts can help the planner generate reasonable sub-tasks to execute for the subsequent planning.

The dynamic sub-skill planner generates the next sub-skill to execute according to the currently planned sub-task. The prompts include the task description, the candidate sub-skill description, the output format, the in-context examples, the state description of the ad hoc agent, the detection information, the abilities and states of teammates, the communication messages, and the current sub-task, as is shown in Fig. 6. The candidate sub-skill description gives the categories of sub-skill that the planner can generate from. Since the sub-skill planner needs to consider the current sub-task to generate the corresponding sub-skill, the prompts also include the current sub-task. We utilize the sub-skill planning prompts based on CoT to inform the LLM to generate the reasonable sub-skill to execute. The designed prompts can help the sub-skill planner generate reasonable sub-skills based on the planned sub-task and the current state.

\subsection{Details of Dataset}
When generating the benchmark dataset, we consider heterogeneous capabilities of teammates and ad hoc agent. In the characteristics tuple of teammates $(\alpha_{nav}, \alpha_{manip}, h, m, step)$, $\alpha_{nav}=1$ denotes that the agent has the navigation ability, otherwise $\alpha_{nav}=0$. $\alpha_{manip}=1$ denotes the agent has the manipulation ability, otherwise $\alpha_{manip}=0$. $h=1$ indicates the height of the agent is high and the height of the camera is around 1.6m, $h=0$ means the height of the agent is low and the height of the camera is around 0.9m. The value of $m$ is a real number ranging from $[0.1,20]$, indicating the maximum weight the object that can be picked up by the agent. The value of $step$ is an integer ranging from $[50,500]$, indicating the maximum steps that the agent can execute. In the tidying-up task, all agents have the navigation ability. When generating teammates, we firstly randomly generate $\alpha_{manip}$ for each teammate and guarantee that at least one teammate has the manipulation ability to complete the task. Then we randomly generate the value of $h$, $m$ and $step$ from their range to form the original team. When generating the capability of the ad hoc agent, we randomly generate $\alpha_{manip}$, $h$, $m$, and $step$ to form the characteristics tuple of it, and the ad hoc agent is likely to have or not have the manipulation ability.

\begin{figure}
	\centering
	\includegraphics[width=3.45in]{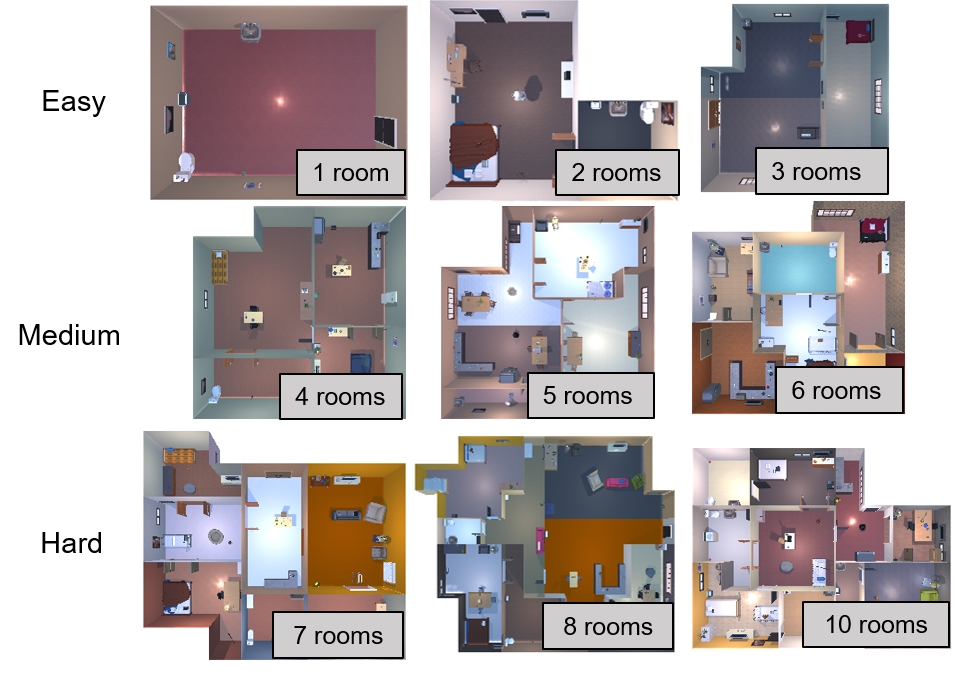}
	\caption{The floorplan samples of the house with different number of rooms in the dataset.}
	\label{fig:sim_room}
\end{figure}

Further, we randomly selected 90 houses from ProcTHOR-10K among which every 10 houses are with 1-8, and 10 rooms respectively (there is no house with 9 rooms in ProcTHOR-10K). We divide all tasks into \textit{Easy}, \textit{Medium} and \textit{Difficult} tasks according the number of rooms in houses. The floorplan samples of houses with different number of rooms are shown in Fig. \ref{fig:sim_room}.

\subsection{Implementation Details}
In the experiments, agents with different capabilities may have different action spaces. The navigation action space is denoted as $I_{nav}=\{MoveAhead,\; \allowbreak
MoveBack, \; \allowbreak
MoveRight, \; \allowbreak
MoveLeft, \; \allowbreak
RotateRight,\; \allowbreak
RotateLeft,\; \allowbreak
LookUp,\; \allowbreak
LookDown,\;
Stop\}$. The manipulation action space is denoted as $I_{mani}=\{PickUp, \; \allowbreak
PutDown, \; \allowbreak
Drop\}$. If $\alpha_{nav}=1$ and $\alpha_{manip}=1$ in the characteristics tuple, the agent can choose low-level action from $I^{(i)} = I_{nav} \cup I_{mani}$ to execute in the sub-skill execution module. If $\alpha_{nav}=1$ and $\alpha_{manip}=0$ in the, the agent only has the navigation ability and can choose low-level action from $I_{nav}$.

In the experiments, the ad hoc policy based on the LLM including \textit{Naive-LLM}, \textit{GPL-LLM}, \textit{CoT} utilize the \textit{gpt-4-1106-preview} to generate corresponding planning results. Our newly proposed model \textit{IRoT-LLM} utilizes the \textit{gpt-4-1106-preview} in the dynamic sub-task planner and sub-skill planner. We set the parameter \textit{Temperature} to be 0.3 in our experiments.

\subsection{Details of the Teammate Policy}

\subsubsection{\textbf{Teammates with Heuristic-Based Policy}}
\label{teamrule}
We define several rules for the collaboration of multiple heterogeneous teammates. Specifically, if no misplaced object is detected by teammates, they will collaboratively explore the whole house based on the frontier-based exploration method in the multi-agent setting. If someone detects a misplaced object, then the agent that has the manipulation ability and is closest to the misplaced objects based on the predicted location information will choose to perform the sub-task to re-place the misplaced object to the reasonable receptacle. Other teammates will continue exploring the house. Whenever a misplaced object is detected, the sub-task of re-place the newly detected misplaced object will be assigned to the teammate which satisfies the above conditions and is not performing the re-place sub-task. If all teammates with the manipulation ability have been assigned the re-place sub-task, it is required to wait for someone completing the executing re-place sub-task and then switch to the new re-place sub-task.

\begin{figure*}
	\centering  
	\includegraphics[width=7in]{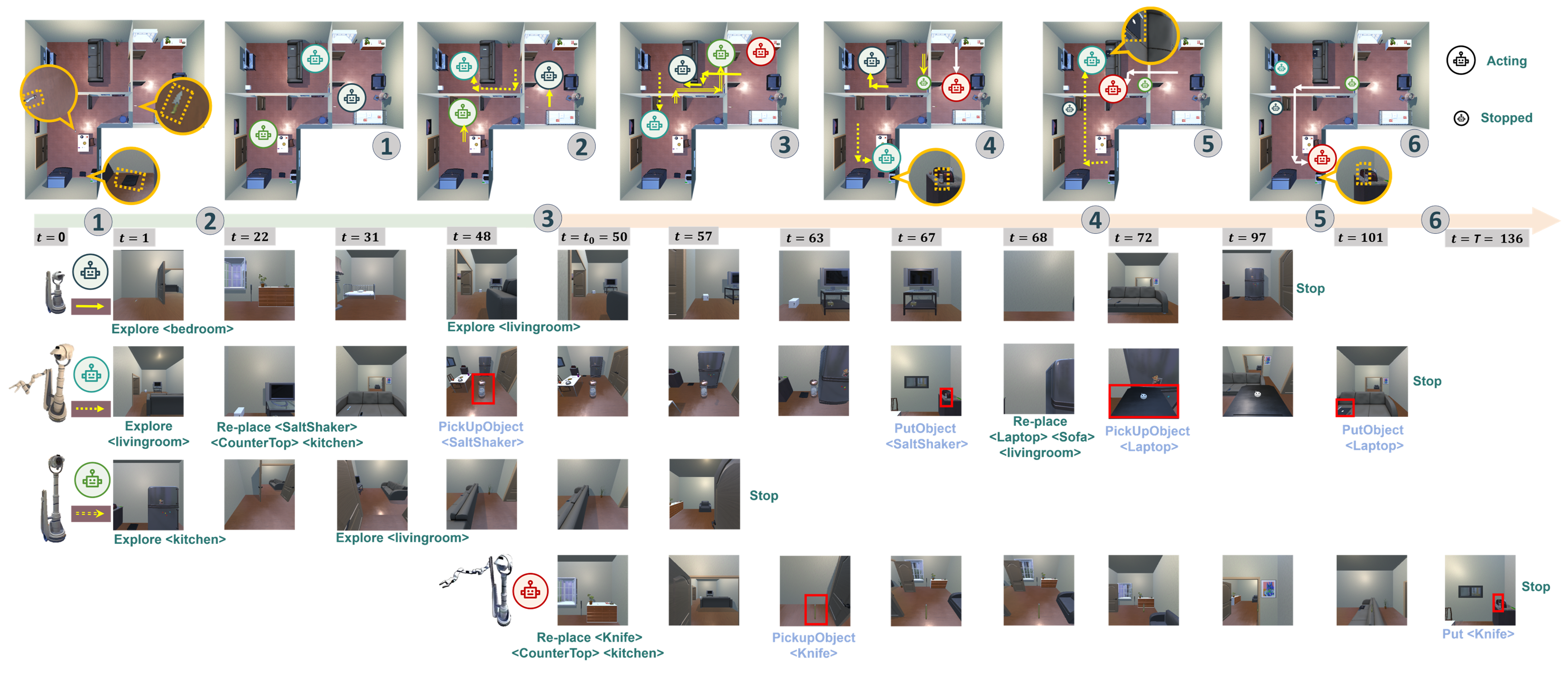}
	\caption{The successful examples of the \textit{IRoT-LLM} ad hoc agent. The icons of different colors mark positions of agents. Yellow lines with different shapes are used to mark the paths between key steps of teammates, and white lines are used to mark the paths of the ad hoc agent. The ad hoc agent joins the original at the 50-th step and help the original team to complete the task.}
	\label{fig:sim_first}
\end{figure*}

\subsubsection{\textbf{Teammates with Learning-based Policy}}
\label{teamlearn}
The learning-based model includes the visual semantic perception, sub-skill decision and the action execution module. The visual semantic perception is similar to that in our proposed model, which obtains the semantic map and the hierarchical scene graph. In the learning-based method, the visual semantic perception module also includes a misplaced object detector consisting of two linear layers. It takes the detected object, its current receptacle and the current room type as the input, and output whether the object is misplaced. We train it in the training scenes of ProcTHOR-10k with the ground truth of the object placement constraints in ProcTHOR in advance. The sub-skill decision module is trained to predict the sub-skill for the agent. It encodes the abilities of the agent, the current location and state of the agent, as well as the spatial relationship of detected objects to obtain the detection embedding, and fuse these embeddings to get the overall state feature. Then it fuses its state feature with others' state features and utilizes the fused feature to predict the type of the next sub-skill. The candidate type of the predicted sub-skill for this learning method include \textit{Explore}, \textit{PickupObject}, \textit{PutObject} and \textit{Stop}. The action execution module executes the predicted sub-skill with the rule-based strategy. We also construct a reasonable receptacle predictor based on the neural network linear layers. It tasks the type of the misplaced object as input and output the choose the reasonable type of the receptacle and room from the candidate list in ProcTHOR. The reasonable receptacle predictor is also pre-trained with the metadata of the object placement constraints in ProcTHOR. When the predicted sub-skill type is \textit{PickupObject}, if an object is detected to be misplaced, then the agent will choose to pickup the corresponding object with the action execution module, otherwise the agent will continuing exploring the scene. If the predicted sub-skill type is \textit{PutObject}, the action execution module will firstly predicts the reasonable location with the reasonable receptacle predictor, and then put the hold object to the predicted location. We utilize the metadata of the task and scene to generate the expert demonstration for the predicted sub-skill for each agent. We train the sub-skill decision module with the imitation learning by the behavior cloning method.

\subsubsection{\textbf{Teammates with the LLM}}
\label{teamLLM}
The structure of the teammates' model based on the LLM consists of the visual semantic perception module, negotiation-based sub-task planner, sub-skill planner, sub-skill execution, and the communication module.

\textbf{Visual Semantic Perception.} The visual perception module is utilized to generate the scene graph and semantic map in the same way as the visual semantic perception module in the proposed ad hoc teamwork framework.


\textbf{Negotiation-based Sub-Task Planner.} The negotiation-based sub-task planner generates the next sub-task based on the negotiation dialog among teammates with the LLM, which include \textit{Explore} $\left\langle ID_{room} \right\rangle$ and \textit{Re-place} $\left\langle ID_{object} \right\rangle$ $\left\langle Type_{receptacle} \right\rangle$ $\left\langle Type_{room} \right\rangle$ in the multi-agent tidying-up task. Due to the different abilities of heterogeneous agents, each agent needs to decide the next sub-task to execute while carefully considering their skills and current states when the misplaced objects are detected. The predicted sub-tasks of each agent need to meet the requirements of their own abilities and avoid planning conflicts among agents. We utilize the reasoning and planning abilities of the LLMs to complete sub-task planning with the negotiation dialog of agents. When the task starts, each agent communicates its ability characteristics to other teammates, and plans the next sub-task based on the CoT approach with the information of teammates' abilities and observations in sequence. Each teammate's abilities, the information of currently detected objects, the historical planning results, and the current predicted sub-task by other teammates who have already finished planning are provided to the LLM-based planner to generate the next sub-task for the corresponding agent. The planner first generates the answer whether it agrees with the already finished planning results by previous agents. If the agent agrees to the finished planning results, it then outputs its planned sub-task to perform. Otherwise, the previous agents will negotiate and re-plan the sub-task again with this feedback information.

\textbf{Dynamic Sub-Skill Planner.} The dynamic sub-skill planner leverages the LLM to predict the next sub-skill with the CoT method, considering their abilities, current and historical states, communication messages as well as the predicted sub-tasks. Each teammate generate the next sub-skill to execute in the similar way as the dynamic sub-skill in the proposed ad hoc teamwork model.

\textbf{Sub-Skill Execution.} The sub-skill execution module executes the predicted sub-skill and updates the newly detected information. We generate policies for each sub-skill in advance and utilize the rule-based methods based on the frontier-based approach to generate low-level actions. 

\textbf{Communication.} Each teammate utilizes the communication module to transmit the key information. At the beginning of the task, each agent broadcasts its ability and detection information through communication messages. Then teammates utilize the extracted ability and state information to plan the next sub-tasks and sub-skills. Similar to the communication module in the proposed ad hoc teamwork framework, teammates transmit key information of detected misplaced objects and candidate receptacles to others, as well as broadcast the state information when finish or switch to new sub-tasks and sub-skills.

\textbf{Adaptive Adjustment to Ad Hoc Agent.} In our ad hoc teamwork task, we consider whether the strategies of teammates adapt to the policy of the ad hoc agent after the ad hoc agent joining the teamwork. Therefore, we also consider utilizing the extra adaptive prompts when planning to explicitly make teammates ponder the situation that an existing agent is broken or a new ad hoc agent is added to the team.

\begin{figure*}
	\centering
	\includegraphics[width=6.45in]{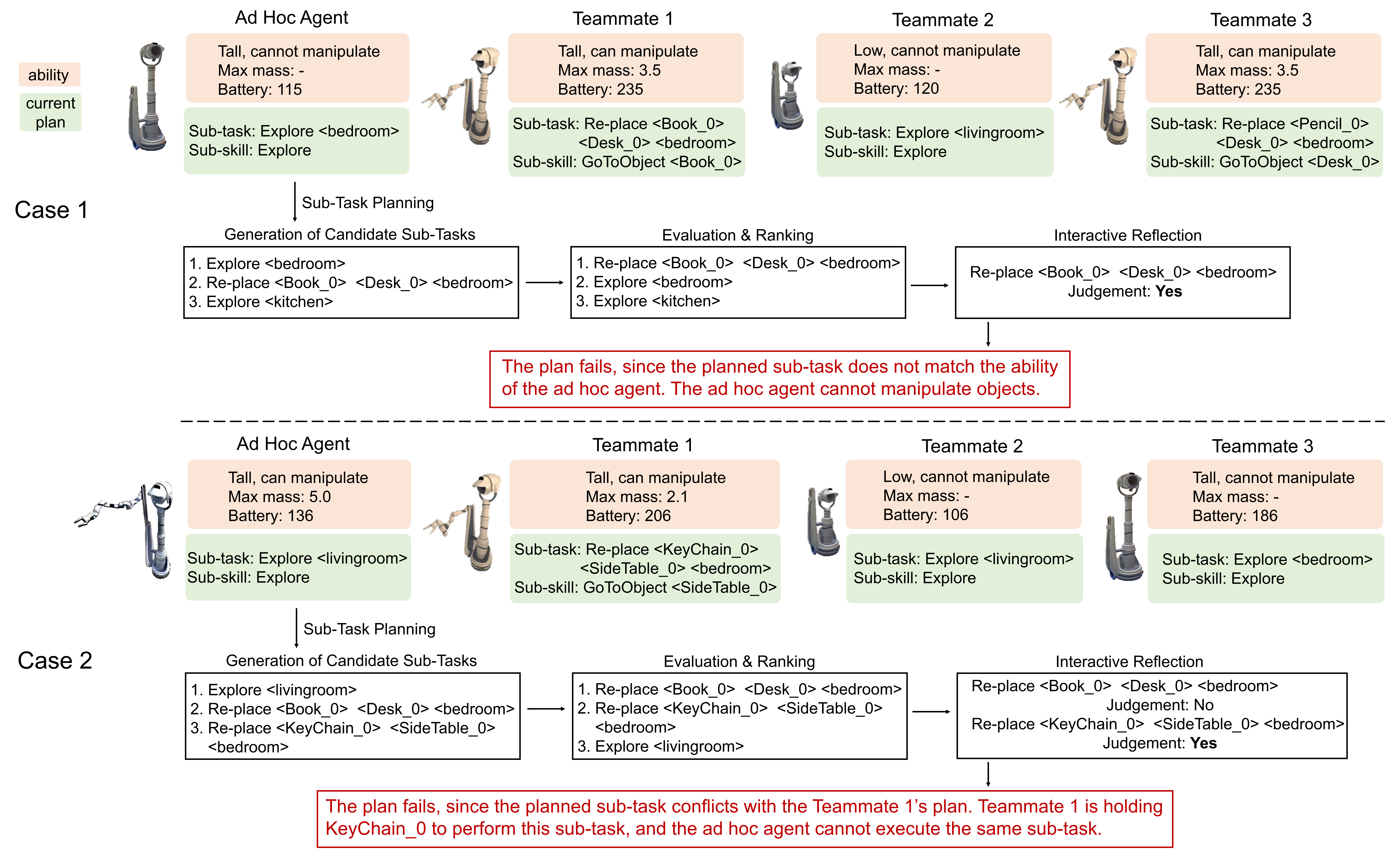}
	\caption{{The failure examples of the sub-task planner of the ad hoc agent. In \textit{Case 1}, the ad hoc agent does not have the manipulation ability, but it plans to execute the \textit{Re-place} sub-task, which does not match the ability of the ad hoc agent. In \textit{Case 2}, the ad hoc agent has the manipulation ability, but its plan conflicts with the current sub-task plan of \textit{Teammate1}.}}
	\label{fig:FailSample}
\end{figure*}

\subsection{Qualitative Performance in Simulation}
We show a successful example of the proposed ad hoc teamwork based on IRoT in the simulation scene with both the first-person view and the top-down view in Fig. \ref{fig:sim_first}. The ad hoc agent with the manipulation abiliy joins the original at the 50-th step. It can help the original team and re-place the misplaced \textit{Knife} to \textit{CounterTop} in the kitchen. The joining of this ad hoc agent improve the efficiency of the original team.

{We also show the failure cases of the dynamic sub-task planner of the ad hoc agent in Fig. \ref{fig:FailSample}. In \textit{Case 1}, the ad hoc agent does not have the manipulation ability. It plans to execute the sub-task \textit{Re-place} $\left\langle Book\_0 \right\rangle$ $\left\langle Desk\_0 \right\rangle$ $\left\langle bedroom \right\rangle$ but the ad hoc agent cannot manipulate the object, The planned sub-task does not match the ability of the ad hoc agent. In \textit{Case 2}, the ad hoc agent has the manipulation ability. It plans to execute the sub-task \textit{Re-place} $\left\langle KeyChain\_0 \right\rangle$ $\left\langle SideTable\_0 \right\rangle$ $\left\langle bedroom \right\rangle$, but \textit{Teammate1} is performing the same sub-task holding the object \textit{KeyChain\_0}. The ad hoc agent cannot execute this sub-task and it conflicts with the current sub-task plan of \textit{Teammate1}.}

\end{document}